\providecommand{\pgfsyspdfmark}[3]{}

\documentclass[11pt,a4paper]{article}
\usepackage[hyperref]{emnlp2020}
\usepackage[utf8]{inputenc} %
\usepackage[T1]{fontenc}    %
\usepackage{times}
\usepackage{latexsym}

\usepackage{microtype}

\aclfinalcopy %
\def\aclpaperid{1220} %

\usepackage{algorithmicx}
\usepackage{algpseudocode}
\usepackage{algorithm}
\usepackage{calc}
\usepackage{nicefrac}
\usepackage{multirow}
\usepackage{booktabs}
\usepackage{natbib}
\usepackage{amsmath,amsthm,amssymb}
\usepackage{framed}
\usepackage{graphicx}
\usepackage{url}            %
\usepackage{amsfonts}       %
\usepackage{enumitem}
\usepackage{xspace}

\newcounter{notecounter}
\newcommand{\enotesoff}{\long\gdef\enote##1##2{}}
\newcommand{\enoteson}{\long\gdef\enote##1##2{{
\stepcounter{notecounter}
{\large\bf \hspace{1cm}\arabic{notecounter} $<<<$ ##1: ##2 $>>>$\hspace{1cm}}}}}
\enoteson

\def\myparagraph#1{\paragraph{#1}}

\def\figref#1{Figure~\ref{fig:#1}}
\def\figlabel#1{\label{fig:#1}\label{p:#1}}

\def\tabref#1{Table~\ref{tab:#1}}
\def\tablabel#1{\label{tab:#1}\label{p:#1}}

\def\secref#1{\S\ref{sec:#1}}
\def\seclabel#1{\label{sec:#1}}
\def\eqref#1{Eq.~\ref{eqn:#1}}

\def\eqlabel#1{\label{eqn:#1}}

\usepackage{multirow}
\usepackage[caption=false]{subfig}
\usepackage{diagbox}

\def\ptl{pretrained language model\xspace}

\def\ptls{pretrained language models\xspace}
\def\Ptls{Pretrained language models\xspace}

\providecommand{\abs}[1]{\left\lvert#1\right\rvert}
\providecommand{\norm}[1]{\left\lVert#1\right\rVert}

\providecommand{\R}{\mathbb{R}} %

\providecommand{\E}{{\mathbb E}}
\providecommand{\E}[1]{{\mathbb E}\left.#1\right. }        %

\providecommand{\tt}{\mathbf{t}}

\providecommand{\ww}{\mathbf{w}}

\providecommand{\mK}{\mathbf{K}}

\providecommand{\mM}{\mathbf{M}}
\providecommand{\mMb}{\mM_\textsf{\tiny bin}}

\providecommand{\mQ}{\mathbf{Q}}

\providecommand{\mV}{\mathbf{V}}
\providecommand{\mW}{\mathbf{W}}
\providecommand{\hmW}{\hat{\mW}}
\providecommand{\mX}{\mathbf{X}}

\providecommand{\mtheta}{\mathbf{\theta}}

\providecommand{\cL}{\mathcal{L}}

\makeatletter
\def\thanks#1{\protected@xdef\@thanks{\@thanks
        \protect\footnotetext{#1}}}
\makeatother

\title{Masking as an Efficient Alternative to Finetuning\\ for Pretrained Language Models}

\author{
    Mengjie Zhao\textsuperscript{†}\textsuperscript{*}\thanks{\ *\ Equal contribution.}, 
    Tao Lin\textsuperscript{‡}\textsuperscript{*},
    Fei Mi\textsuperscript{‡}, 
    Martin Jaggi\textsuperscript{‡},
    Hinrich Sch\"{u}tze\textsuperscript{†}\\
    \textsuperscript{†} LMU Munich, Germany \ 
    \textsuperscript{‡} EPFL, Switzerland\\
    {\tt mzhao@cis.lmu.de, \{tao.lin, fei.mi, martin.jaggi\}@epfl.ch}
  }

\date{}

\begin{document}
\maketitle

\enotesoff

\begin{abstract}
	We present an efficient method of utilizing pretrained language models,
	where we learn selective binary masks for pretrained weights in lieu of
	modifying them through finetuning.
        Extensive evaluations of masking BERT, RoBERTa, and DistilBERT
	on eleven diverse NLP tasks show that our masking scheme yields performance
	comparable to finetuning, yet has a much smaller memory
	footprint when several tasks need to be inferred.
	Intrinsic evaluations show that representations computed by our binary masked
	language models encode information necessary for solving downstream tasks.
	Analyzing the loss landscape, we show
	that masking and finetuning  produce models that
	reside in minima that can be connected by a line segment with nearly
	constant test accuracy. This confirms that masking can be utilized as an
	efficient alternative to finetuning.
\end{abstract}

\section{Introduction}
Finetuning a large \ptl like BERT \citep{devlin-etal-2019-bert}, RoBERTa
\citep{liu2019roberta}, and XLNet \citep{yang2019xlnet} often yields competitive
or even state-of-the-art results on NLP benchmarks \citep{wang-etal-2018-glue,
	wang2019superglue}. Given an NLP task,
standard finetuning stacks a linear
layer on top of the \ptl and then updates all parameters using mini-batch SGD.
Various aspects like brittleness \citep{dodge2020finetuning}
and adaptiveness \citep{peters-etal-2019-tune} of this two-stage transfer
learning NLP paradigm \citep{NIPS2015_5949,howard-ruder-2018-universal} have
been studied.

Despite the simplicity and  impressive performance of
finetuning, the prohibitively large number of parameters to be finetuned,
e.g., 340 million in BERT-large, is a major obstacle to
wider deployment of these models. The large memory footprint
of finetuned models
becomes more
prominent when multiple tasks need to be solved -- several copies
of the millions of finetuned parameters have to be saved for inference.

Recent work~\citep{gaier2019weight,zhou2019deconstructing}
points out the potential of searching neural architectures within a fixed model,
as an alternative to optimizing the model weights for downstream tasks.
Inspired by these results, we present \emph{masking}, a simple yet efficient scheme for utilizing \ptls.
Instead of directly updating the pretrained parameters,
we propose to \emph{select} weights important to downstream NLP tasks
while \emph{discarding} irrelevant ones.
The selection
mechanism consists of a set of binary masks, one learned per downstream task
through end-to-end training.

We show that masking, when being applied to \ptls like BERT,
RoBERTa, and DistilBERT \citep{sanh2019distilbert}, achieves performance comparable to finetuning
in tasks like part-of-speech tagging, named-entity recognition,
sequence classification, and reading comprehension.
This is surprising in that
a simple subselection mechanism
that does not change any weights
is competitive with a
training regime -- finetuning -- that can change the value
of every single weight.
We conduct detailed analyses
revealing important factors and possible reasons for the desirable performance of masking.

Masking  is parameter-efficient: only a set of 1-bit binary masks
needs to be saved per task after training, instead of all  32-bit float
parameters in finetuning.
This small memory footprint enables deploying \ptls for solving
multiple tasks on edge devices. The compactness of
masking also naturally allows parameter-efficient ensembles of \ptls.

Our \textbf{contributions}:
(i) We introduce \emph{masking}, a new scheme for utilizing \ptls\ by learning
selective masks for pretrained weights, as an efficient alternative to finetuning.
We show that masking is applicable to models like
BERT/RoBERTa/DistilBERT, and produces performance on par with finetuning.
(ii) We carry out extensive empirical analysis of masking,
shedding light on factors critical for achieving good performance
on eleven diverse NLP tasks.
(iii) We study the binary masked language models'
loss landscape and language representations,
revealing potential reasons why  masking
has task performance comparable to finetuning.

\section{Related Work}
\myparagraph{Two-stage NLP paradigm.} \Ptls
\citep{peters-etal-2018-deep,devlin-etal-2019-bert,
	liu2019roberta,yang2019xlnet,radford2019language} advance NLP with
contextualized representation of words. Finetuning a \ptl
\citep{NIPS2015_5949,howard-ruder-2018-universal} often delivers competitive
performance
partly because
pretraining leads to a better initialization across various downstream tasks than
training from
scratch \citep{hao-etal-2019-visualizing}. However, finetuning on
individual NLP tasks is not parameter-efficient. Each finetuned model,
typically consisting of hundreds of millions of floating point parameters,
needs to be saved individually.
\citet{pmlr-v97-stickland19a}
use projected attention layers with multi-task learning to improve efficiency
of finetuning BERT. \citet{pmlr-v97-houlsby19a} insert adapter modules to BERT
to improve memory efficiency.
The inserted modules alter the forward pass of BERT,
hence need to be carefully initialized to be close to identity.

We propose to directly pick parameters appropriate to a
downstream task, by learning selective binary masks via end-to-end training.
Keeping the pretrained parameters untouched, we solve several downstream NLP
tasks with minimal overhead.

\myparagraph{Binary networks and network pruning.}
Binary masks can be trained using
the ``straight-through estimator''
\citep{bengio2013estimating,hintonestimator}.
\citet{NIPS2016_6573}, \citet{rastegari2016xnor}, \citet{hubara2017quantized},
\emph{inter alia},
apply this technique to train efficient binarized neural networks.
We use this estimator to train selective masks
for pretrained language model parameters.

Investigating the lottery ticket hypothesis \citep{frankle2018lottery} of
network pruning~\citep{han2015learning,he2018soft,liu2018rethinking,lee2018snip,Lin2020Dynamic},
\citet{zhou2019deconstructing} find that applying binary masks to a neural
network is a form of training the network.
\citet{gaier2019weight} propose to search neural architectures for reinforcement
learning and image classification
tasks, without any explicit weight training.
This work inspires our  masking scheme
(which can be interpreted as implicit neural architecture search~\citep{liu2018rethinking}):
applying the masks to a \ptl is similar to finetuning, yet
is much more parameter-efficient.

Perhaps the closest work, \citet{Mallya_2018_ECCV} apply binary masks to CNNs
and achieve good performance in computer vision.
We learn selective binary masks for \ptls in NLP and shed
light on factors important for obtaining good performance.
\citet{Mallya_2018_ECCV}  explicitly update weights in a
task-specific classifier layer. In contrast,
we show that end-to-end learning of selective masks, consistently for
both the \ptl and a randomly initialized classifier layer,
achieves good performance.
\citet{radiyadixit2020fine}
investigate finetuning of BERT by employing a number of techniques,
including what they call sparsification, a method similar to
masking. Their focus is analysis of finetuning BERT whereas our goal
is to provide an efficient alternative to finetuning.

\section{Method}
\subsection{Background on Transformer and finetuning}
The encoder of the Transformer
architecture \citep{vaswani2017attention} is ubiquitously used when
pretraining large language models.  We briefly review its architecture
and then present our masking scheme.  Taking BERT-base as an example,
each one of the 12 transformer blocks consists of (i) four linear
layers\footnote{We omit the bias terms for brevity.} $\mW_K$, $\mW_Q$,
$\mW_V$, and $\mW_{AO}$ for computing and outputting the self
attention among input wordpieces \citep{wu2016googles}.
(ii) two linear layers $\mW_I$ and $\mW_O$ feeding forward the word
representations to the next transformer block.

More concretely, consider an input sentence $\mX \in \R^{N \times d}$
where $N$ is the maximum sentence length and $d$ is the hidden dimension size.
$\mW_K$, $\mW_Q$, and $\mW_V$ are used to compute transformations of $\mX$:
\begin{align*}
	\mK = \mX \mW_K, \mQ = \mX \mW_Q, \mV = \mX \mW_V, 
\end{align*}
and the self attention of $\mX$ is computed as:
\begin{align*}
	\text{Attention($\mK, \mQ, \mV$)} = \text{softmax}(\frac{\mQ \mK^{T}}{\sqrt{d}}) \mV.
\end{align*}

The attention is then transformed by $\mW_{AO}$, and subsequently
fed forward by $\mW_I$ and $\mW_O$ to the next transformer block.

When finetuning on a downstream task like sequence classification, a
linear classifier layer $\mW_T$, projecting from the hidden dimension to the output dimension,
is randomly initialized.
Next, $\mW_T$ is stacked on top of a pretrained linear layer
$\mW_P$ (the \emph{pooler layer}).
All parameters are then updated to minimize the task loss such as cross-entropy.

\subsection{Learning the mask}
\seclabel{masklearning}
Given a \ptl, we
do not finetune, i.e., we do
not update the pretrained parameters.
Instead, we \emph{select} a
subset of the pretrained parameters that is critical
to a downstream task while \emph{discarding} irrelevant ones with binary masks.
We associate each linear layer
$\mW^{l}$ $\in$ $\{\mW^{l}_K, \mW^{l}_Q, \mW^{l}_V, \mW^{l}_{AO}, \mW^{l}_I, \mW^{l}_O\}$
of the $l$-th transformer block
with a real-valued matrix $\mM^{l}$
that is randomly initialized
from a uniform distribution and has the same size as $\mW^{l}$. We then pass
$\mM^{l}$ through an element-wise thresholding function
\citep{NIPS2016_6573,Mallya_2018_ECCV}, i.e., a binarizer,
to obtain a binary mask $\mMb^{l}$ for $\mW^{l}$:
\begin{align} \eqlabel{threshold_func}
	\vspace{-2em}
	\textstyle
	(m_\textsf{\tiny bin}^{l})_{i, j}
	= \left\{ \begin{array}{ll}
		1 & \textrm{if $m^{l}_{i, j} \geq \tau$} \\
		0 & \textrm{otherwise}
	\end{array} \right. \,,
	\vspace{-2em}
\end{align}
where $m^{l}_{i, j} \in \mM^{l}$,  $i, j$ indicate the
coordinates of the 2-D linear layer and
$\tau$ is a global thresholding hyperparameter.

In each forward pass of training,
the binary mask $\mMb^{l}$ (derived from $\mM^l$ via \eqref{threshold_func})
selects weights
in a pretrained linear layer $\mW^{l}$ by Hadamard product:
\begin{align*}
	\vspace{-2em}
	\textstyle
	\hmW^{l} := \mW^{l} \odot \mMb^{l} \,.
	\vspace{-2em}
\end{align*}

In the corresponding backward pass of training, with the associated loss function
$\cL$,
we cannot backpropagate through
the binarizer, since \eqref{threshold_func} is a hard thresholding
operation and the gradient with respect to $\mM^{l}$ is
zero almost everywhere.
Similar to the treatment\footnote{
	\citet{bengio2013estimating,NIPS2016_6573} describe it as the ``straight-through estimator'',
	and \citet{Lin2020Dynamic} provide convergence guarantee with error feedback interpretation.
} in~\citet{bengio2013estimating,NIPS2016_6573,Lin2020Dynamic},
we use $\frac{\partial \cL ( \hmW^{l} )}{\partial \mMb^{l}}$
as a noisy estimator of $\frac{\partial \cL ( \hmW^{l} )}{\partial \mM^{l}}$
to update $\mM^{l}$, i.e.:
\begin{align} \eqlabel{mask_update}
	\vspace{-2em}
	\textstyle
	\mM^{l} \gets \mM^{l} - \eta \frac{\partial \cL ( \hmW^{l} )}{\partial \mMb^{l}} \,,
	\vspace{-2em}
\end{align}
where $\eta$ refers to the step size. Hence, the whole structure can be trained
end-to-end.

We learn a set of binary masks for an NLP task as follows.
Recall that each linear layer
$\mW^{l}$ is associated with a $\mM^{l}$ to obtain a
masked linear layer $\hmW^l$ through \eqref{threshold_func}.
We randomly initialize
an additional linear layer with an associated $\mM^{l}$
and stack
it on top of the \ptl.
We
then update each $\mM^{l}$ through \eqref{mask_update} with the task objective during training.

After training, we pass each $\mM^{l}$ through the binarizer to obtain
$\mMb^{l}$, which is then saved for future inference. Since $\mMb^{l}$ is binary,
it takes only $\approx$ 3\% of the memory compared to saving the 32-bit float parameters
in a finetuned model. Also, we will show
that
many layers -- in particular the embedding layer --
do not have to be masked.
This further reduces memory consumption of masking.

\subsection{Configuration of masking}
Our masking scheme is motivated by the observation:
the pretrained weights form a good initialization
\cite{hao-etal-2019-visualizing}, yet a few steps of adaptation are still
needed to produce competitive performance for a specific task.
However, not every pretrained parameter is necessary for achieving reasonable performance, as
suggested by the field of neural network pruning
\citep{NIPS1989_250,NIPS1992_647,NIPS2015_5784}.
We now investigate two configuration choices that affect how
many parameters are ``eligible''
for masking.

\myparagraph{Initial sparsity of $\mMb^{l}$.}
As we randomly initialize our masks from uniform distributions,
the sparsity of the binary mask $\mMb^{l}$ in the mask initialization phase
controls how many pretrained
parameters in a layer $\mW^{l}$
are assumed to be irrelevant to the downstream task.
Different initial sparsity rates entail different optimization behaviors.

It is crucial to better understand
how the initial sparsity of a mask impacts the training dynamics and final model performance,
so as to generalize our masking scheme to broader domains and tasks.
In \secref{initsparsity}, we investigate this aspect in detail.
In practice, we fix $\tau$ in \eqref{threshold_func} while adjusting the uniform distribution to achieve a target initial sparsity.

\myparagraph{Which layers to mask.} Different layers
of \ptls capture distinct aspects of a language during pretraining, e.g.,
\citet{tenney-etal-2019-bert} find that information on
part-of-speech tagging,
parsing, named-entity recognition, semantic roles, and coreference is encoded on progressively higher
layers of BERT.
It is hard
to know a priori  which types of NLP tasks have to be
addressed in the future, making
it non-trivial to decide layers to mask. We study this factor in
\secref{maskwhichlayers}.

We do not learn a mask for the lowest embedding layer, i.e., the
uncontextualized wordpiece embeddings are completely
``selected'', for all
tasks. The motivation is two-fold.
(i) The embedding layer weights
take up a large part, e.g., almost 21\% (23m/109m) in BERT-base-uncased,
of the total number of parameters. Not having to learn a selective mask for this layer
reduces memory consumption. (ii) Pretraining has
effectively encoded context-independent general meanings of words in the embedding
layer~\citep{zhao2020quantifying}.
Hence, learning a selective mask for this layer is unnecessary.
Also, we do not learn masks for biases and layer
normalization parameters as we did not observe
a positive effect on
performance.

\section{Datasets and Setup}
\seclabel{datasetup}
\textbf{Datasets.}
We present results for masking BERT, RoBERTa, and DistilBERT in part-of-speech tagging, named-entity
recognition, sequence classification, and reading comprehension.

We experiment with \textbf{part-of-speech tagging} (POS) on Penn Treebank
\citep{marcus-etal-1993-building}, using \citet{collins-2002-discriminative}'s
train/dev/test split. For \textbf{named-entity recognition} (NER), we conduct experiments
on the CoNLL-2003 NER shared task \citep{tjong-kim-sang-de-meulder-2003-introduction}.

For \textbf{sequence classification},
the following GLUE tasks \citep{wang-etal-2018-glue} are evaluated:
Stanford Sentiment Treebank (SST2) \citep{socher-etal-2013-recursive}, Microsoft
Research Paraphrase Corpus (MRPC) \citep{dolan2005automatically}, Corpus of
Linguistic Acceptability (CoLA) \citep{Warstadt_2019}, Recognizing Textual
Entailment (RTE) \citep{dagan2005pascal}, and Question Natural Language Inference (QNLI)
\citep{rajpurkar-etal-2016-squad}.

In addition, we experiment on sequence classification datasets
that have publicly available test sets:
the 6-class question classification dataset TREC \citep{trec6},
the 4-class news classification dataset AG News (AG) \citep{NIPS2015_5782},
and the
binary Twitter sentiment classification task SemEval-2016 4B (SEM) \citep{nakov-etal-2016-semeval}.

We experiment with \textbf{reading comprehension} on SWAG
\citep{zellers-etal-2018-swag} using the official data splits.
We report Matthew's correlation coefficient (MCC) for CoLA, micro-F1 for NER,
and accuracy for the other tasks.

\textbf{Setup.}
Due to resource limitations and in the spirit of environmental responsibility
\citep{strubell-etal-2019-energy,schwartz2019green}, we conduct our experiments on the base
models: BERT-base-uncased, RoBERTa-base, and DistilBERT-base-uncased.
Thus, the BERT/RoBERTa models we use have 12 transformer blocks (0--11 indexed) producing 768-dimension vectors;
the DistilBERT model we use has the same dimension but contains 6 transformer blocks (0--5 indexed).
We implement
our models in PyTorch~\citep{paszke2019pytorch} with the HuggingFace framework~\citep{Wolf2019HuggingFacesTS}.

Throughout all experiments, we limit the maximum length of a sentence (pair) to
be 128 after wordpiece tokenization. Following \citet{devlin-etal-2019-bert}, we
use the Adam \cite{kingma2014adam} optimizer of which the learning rate is a
hyperparameter while the other parameters remain default.
We carefully tune the learning rate for each setup:
the tuning procedure ensures that the best
learning rate does not lie on the border of our search grid,
otherwise we extend the grid accordingly.
The initial grid is \{1e-5, 3e-5, 5e-5, 7e-5, 9e-5\}.

For sequence classification  and reading comprehension,
we
use \texttt{[CLS]} as the
representation of the
sentence (pair).
Following \citet{devlin-etal-2019-bert}, we  formulate  NER as a tagging task and
use a linear output layer, instead of a conditional random field layer.
For POS and NER experiments,
the representation of a tokenized word
is its last wordpiece
\cite{liu-etal-2019-linguistic,he2019establishing}.
Note that a 128 maximum length of a sentence for POS and
NER means that some word-tag annotations need to be excluded.
Appendix \secref{checklist} shows our reproducibility checklist containing
more implementation and preprocessing details.

\section{Experiments}
\seclabel{experimentsec}
\subsection{Initial sparsity of binary masks}
\seclabel{initsparsity}
We first investigate how initial sparsity percentage (i.e., fraction of zeros) of the binary mask $\mMb^{l}$ influences
performance of a binary masked language model on downstream tasks. We experiment on four tasks,
with initial sparsities in \{1\%, 3\%, 5\%, 10\%, 15\%, 20\%, \ldots, 95\%\}.
All other hyperparameters are controlled: learning rate is fixed
to 5e-5; batch size is 32 for relatively small datasets
(RTE, MRPC, and CoLA) and 128 for SST2.
Each experiment is repeated four times with different random seeds \{1, 2, 3, 4\}. In this experiment, all transformer blocks,
the pooler layer, and the classifier layer are masked.

\begin{figure}[t]
	\centering
	\includegraphics[width=.85\columnwidth]{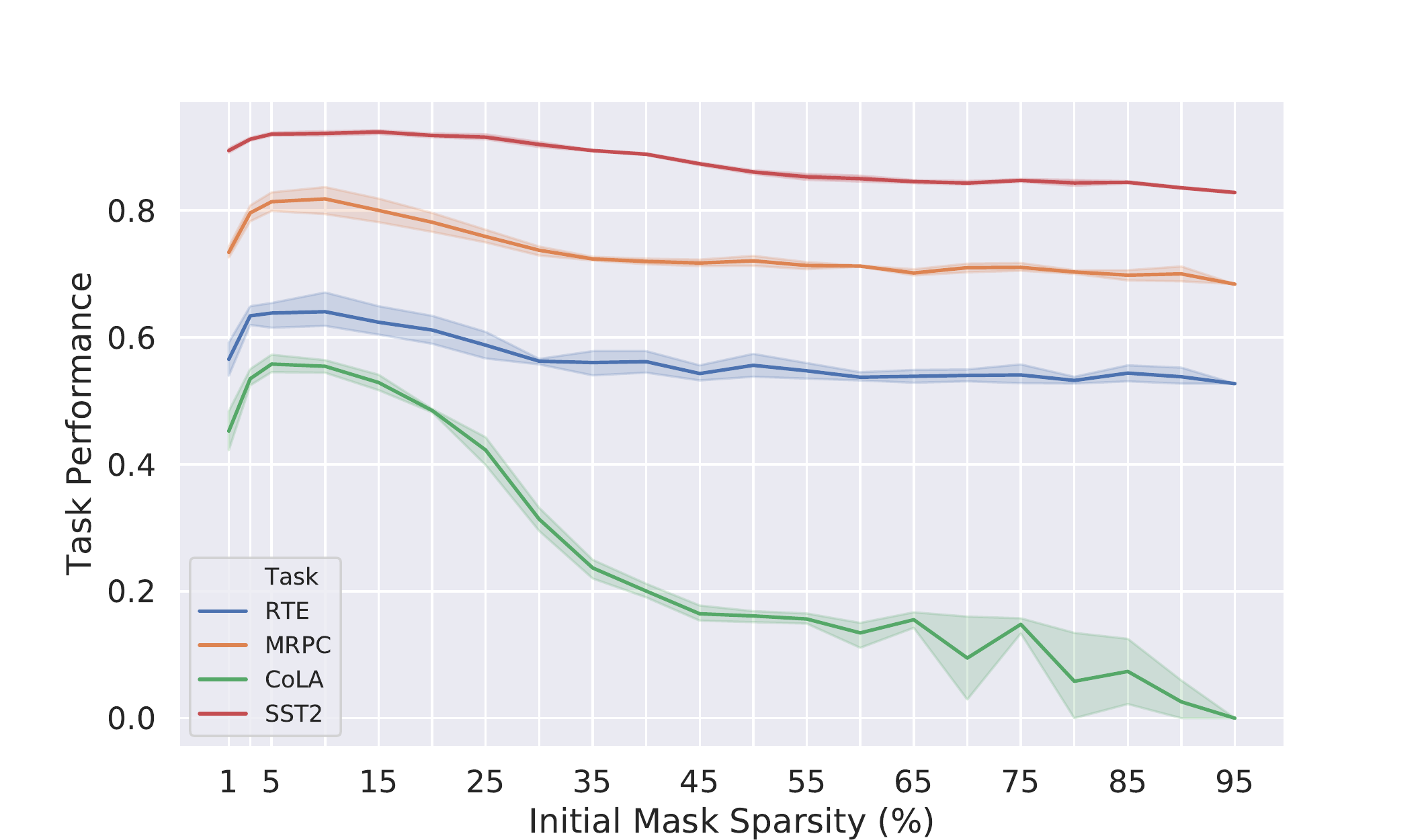}
	\caption{
		Dev set performance of masking BERT when selecting different amounts of
		pretrained parameters.
	}
	\figlabel{initsparsity}
\end{figure}

\figref{initsparsity} shows that
masking  achieves decent performance without
hyperparameter search.
Specifically, (i) a large initial sparsity removing most
pretrained parameters, e.g., 95\%, leads to bad performance
for the four tasks. This is due to the fact that the pretrained knowledge is largely discarded.
(ii) Gradually decreasing the
initial sparsity improves  task
performance. Generally, an initial sparsity in
3\% $\sim$ 10\%
yields reasonable results across tasks.
Large datasets like SST2 are less sensitive than small datasets like RTE.
(iii) Selecting almost all pretrained parameters, e.g.,
1\% sparsity, hurts task performance.
Recall that a pretrained model needs to be adapted to a downstream task;
masking  achieves adaptation by learning
selective masks -- preserving too many pretrained parameters in initialization
impedes the optimization.

\subsection{Layer-wise behaviors}
\seclabel{maskwhichlayers}
Neural network layers present heterogeneous characteristics
\citep{zhang2019all} when being applied to tasks. For example, syntactic
information is better represented at lower layers while semantic information is
captured at higher layers in
ELMo \citep{peters-etal-2018-deep}.
As a result, simply masking all transformer blocks (as in
\secref{initsparsity}) may not be ideal.

\begin{figure*}[h]
	\centering
	\subfloat{
		\includegraphics[width=0.5\columnwidth]{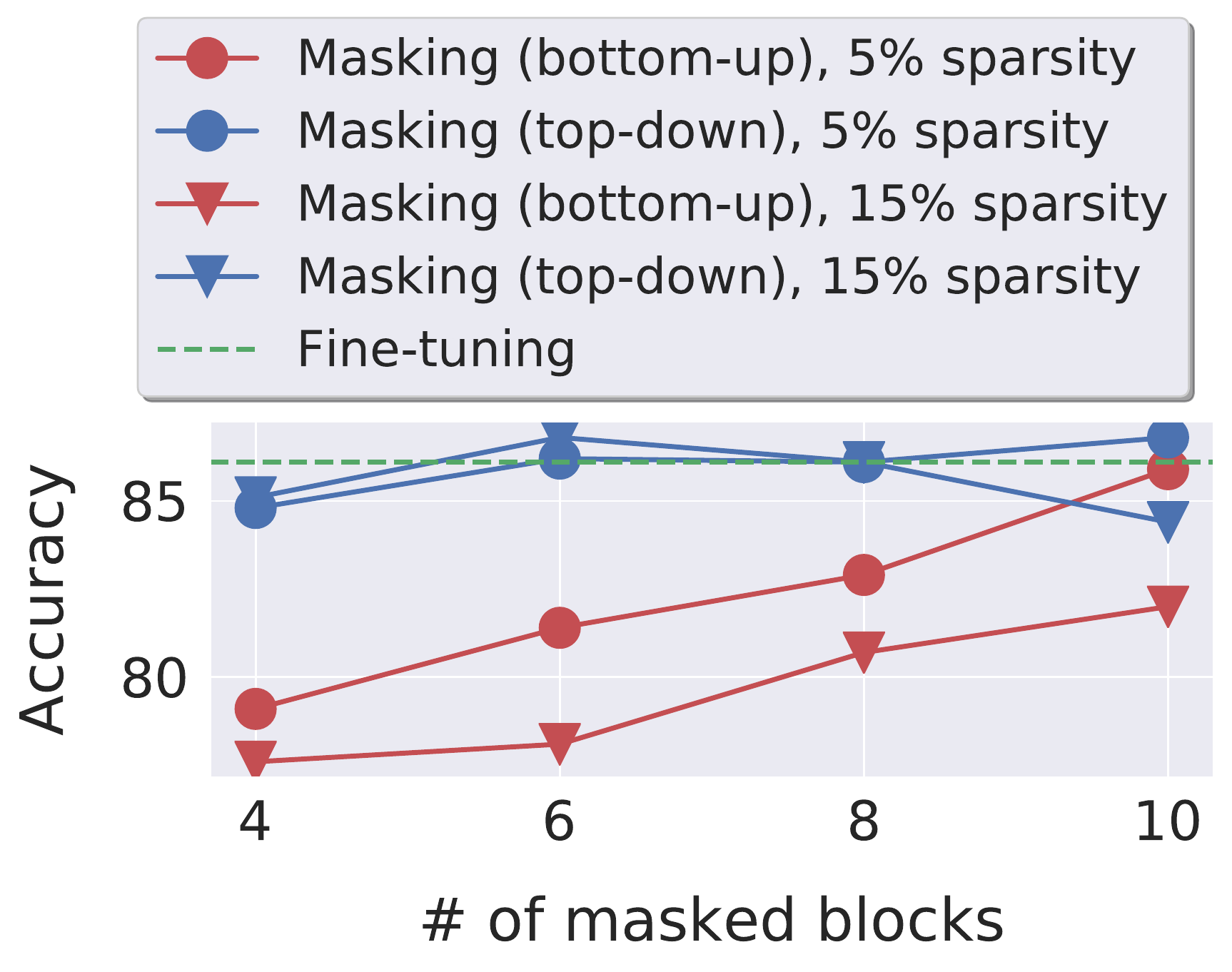}
	}
	\hfill
	\subfloat{
		\includegraphics[width=0.52\columnwidth]{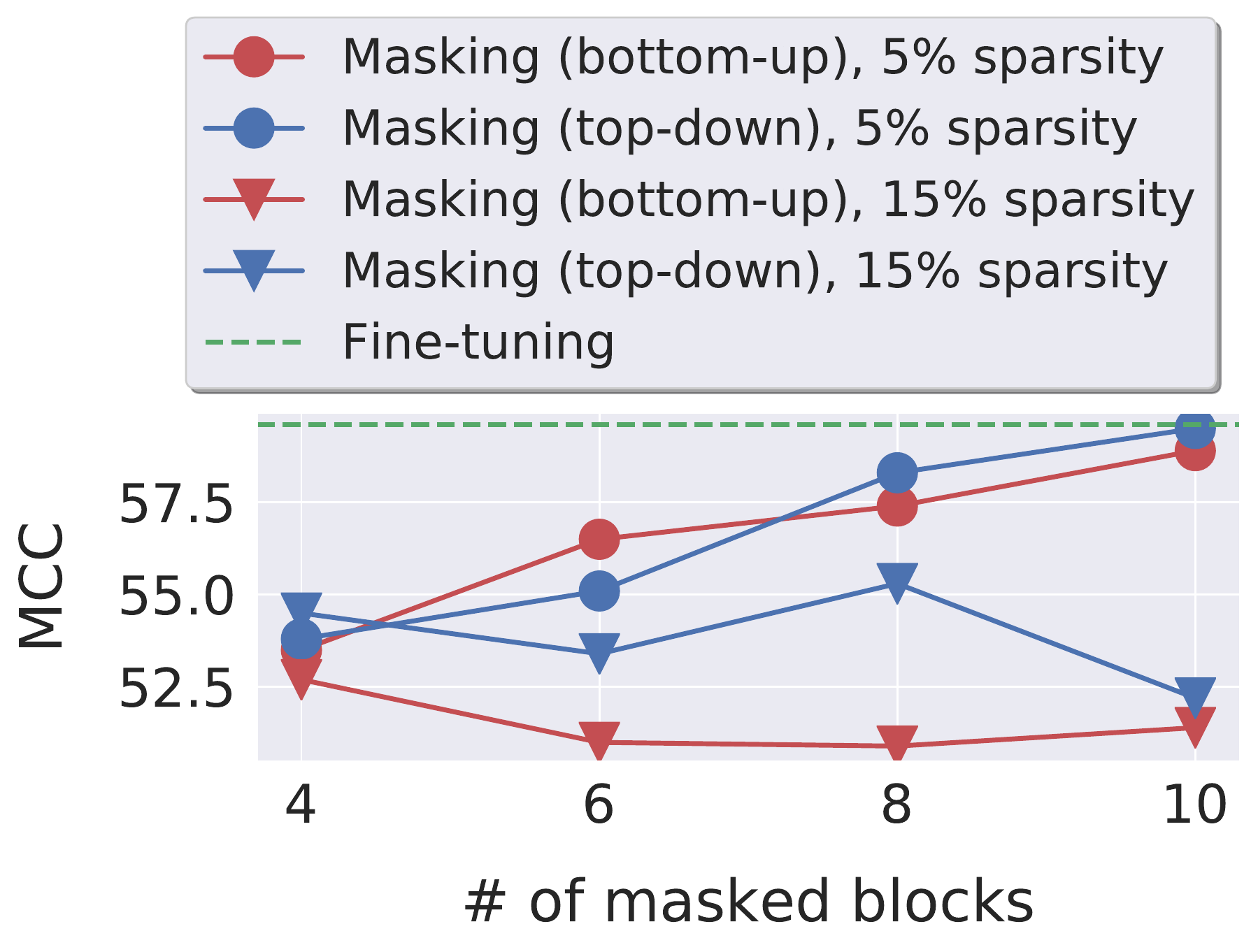}
	}
	\hfill
	\subfloat{
		\includegraphics[width=0.5\columnwidth]{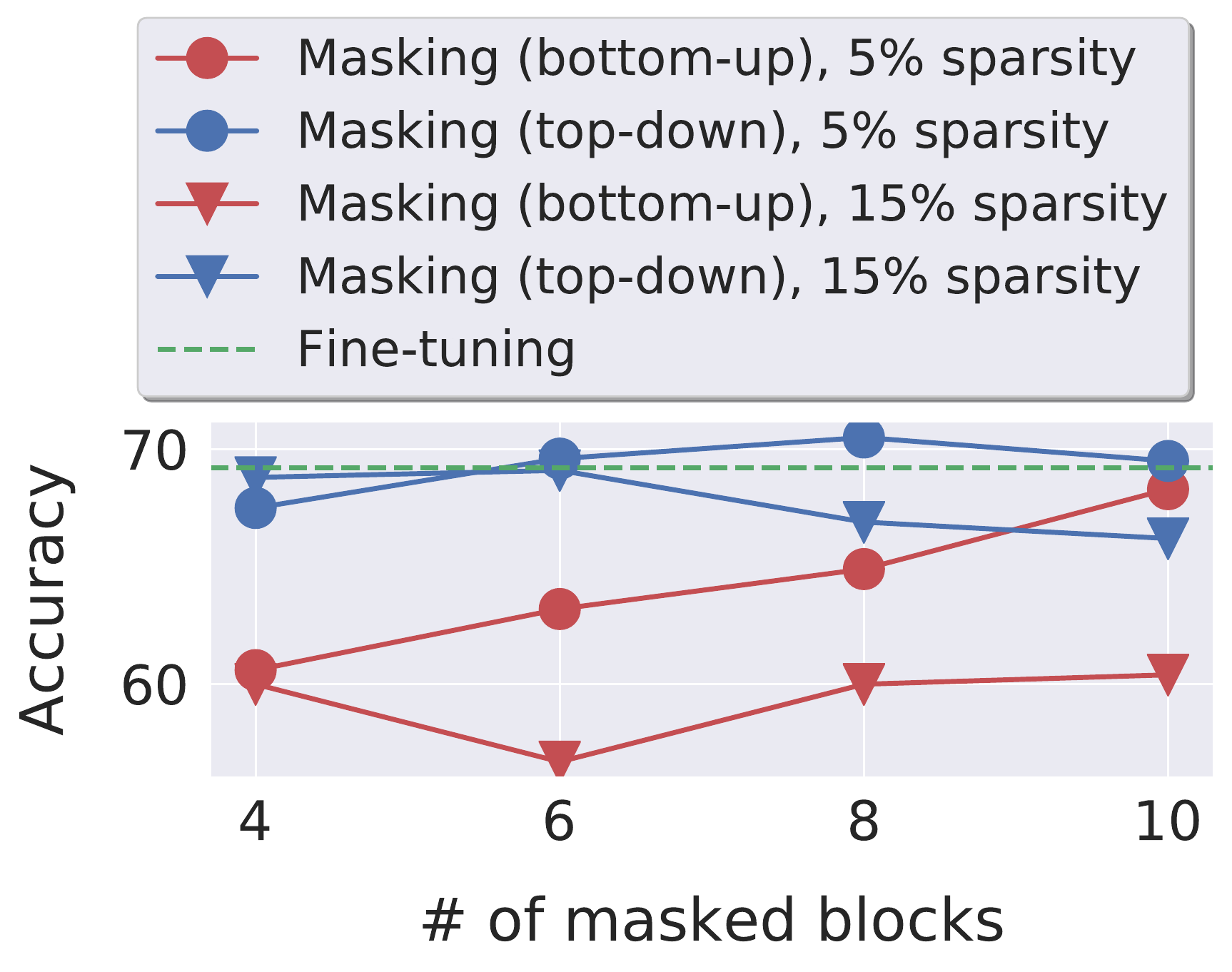}
	}
	\caption{
		The impact of masking different transformer
		blocks of BERT for MRPC (left), CoLA (middle), and
		RTE (right).
		The number of masked blocks is shown on the x-axis;
		that number is either masked ``bottom-up'' or ``top-down''.
		More precisely, a bottom-up setup (red) masking 4 blocks means
		we mask the transformer blocks $\{ 0,1,2,3 \}$; a top-down setup (blue) masking 4 blocks means
		we mask the transformer blocks $\{ 8,9,10,11 \}$.
		$\mW_{P}$ and $\mW_{T}$ are always masked.
	}
	\figlabel{different_layers}
\end{figure*}

\begin{table*}[!ht]
	\centering
	\renewcommand{\arraystretch}{1.2}
	\resizebox{\textwidth}{!}{
		\begin{tabular}{cc|cccccccc|cc|c|}
			\cline{3-13}
			                                                  &            & MRPC           & SST2           & CoLA           & RTE            & QNLI           & SEM            & TREC           & AG             & POS            & NER            & SWAG           \\
			                                                  &            & 3.5k           & 67k            & 8.5k           & 2.5k           & 108k           & 4.3k           & 4.9k           & 96k            & 38k            & 15k            & 113k           \\ \hline
			\multicolumn{1}{|c|}{\multirow{2}{*}{BERT}}       & Finetuning & 86.1 $\pm$ 0.8 & 93.3 $\pm$ 0.2 & 59.6 $\pm$ 0.8 & 69.2 $\pm$ 2.7 & 91.0 $\pm$ 0.6 & 86.6 $\pm$ 0.3 & 96.4 $\pm$ 0.2 & 94.4 $\pm$ 0.1 & 97.7 $\pm$ 0.0 & 94.6 $\pm$ 0.2 & 80.9 $\pm$ 1.7 \\
			\multicolumn{1}{|c|}{}                            & Masking    & 86.8 $\pm$ 1.1 & 93.2 $\pm$ 0.5 & 59.5 $\pm$ 0.1 & 69.5 $\pm$ 3.0 & 91.3 $\pm$ 0.4 & 85.9 $\pm$ 0.5 & 96.0 $\pm$ 0.4 & 94.2 $\pm$ 0.0 & 97.7 $\pm$ 0.0 & 94.5 $\pm$ 0.1 & 80.3 $\pm$ 0.1 \\ \hline
			\multicolumn{1}{|c|}{\multirow{2}{*}{RoBERTa}}    & Finetuning & 89.8 $\pm$ 0.5 & 95.0 $\pm$ 0.3 & 62.1 $\pm$ 1.7 & 78.2 $\pm$ 1.1 & 92.9 $\pm$ 0.2 & 90.2 $\pm$ 0.5 & 96.2 $\pm$ 0.4 & 94.7 $\pm$ 0.0 & 98.1 $\pm$ 0.0 & 94.9 $\pm$ 0.1 & 83.4 $\pm$ 0.8 \\
			\multicolumn{1}{|c|}{}                            & Masking    & 88.5 $\pm$ 1.1 & 94.5 $\pm$ 0.3 & 60.3 $\pm$ 1.3 & 69.2 $\pm$ 2.1 & 92.4 $\pm$ 0.1 & 90.1 $\pm$ 0.1 & 95.9 $\pm$ 0.5 & 94.5 $\pm$ 0.1 & 98.0 $\pm$ 0.0 & 93.9 $\pm$ 0.1 & 82.1 $\pm$ 0.2 \\ \hline
			\multicolumn{1}{|c|}{\multirow{2}{*}{DistilBERT}} & Finetuning & 85.4 $\pm$ 0.5 & 91.6 $\pm$ 0.4 & 55.1 $\pm$ 0.3 & 62.2 $\pm$ 3.0 & 89.0 $\pm$ 0.8 & 85.9 $\pm$ 0.2 & 95.7 $\pm$ 0.6 & 94.2 $\pm$ 0.1 & 97.6 $\pm$ 0.0 & 94.1 $\pm$ 0.1 & 72.5 $\pm$ 0.2 \\
			\multicolumn{1}{|c|}{}                            & Masking    & 86.0 $\pm$ 0.3 & 91.3 $\pm$ 0.3 & 53.1 $\pm$ 0.7 & 61.6 $\pm$ 1.5 & 89.2 $\pm$ 0.2 & 86.6 $\pm$ 0.6 & 95.9 $\pm$ 0.6 & 94.2 $\pm$ 0.1 & 97.6 $\pm$ 0.0 & 94.1 $\pm$ 0.2 & 71.0 $\pm$ 0.0 \\ \hline
		\end{tabular}}
	\caption{
		Dev set task performances (\%) of masking and finetuning.
		Each experiment is repeated four times with different random seeds and we report mean and standard deviation.
		Numbers below dataset name (second row) are the size of training set. For POS and NER, we report the number of sentences.
	}
	\tablabel{devperf}
\end{table*}

We investigate the task performance when applying the masks to different BERT
layers. \figref{different_layers} presents the optimal task performance when
masking only a subset of BERT's transformer blocks  on MRPC, CoLA, and RTE.
Different amounts and indices of transformer blocks are masked:
``bottom-up'' and ``top-down'' indicate to mask the targeted amount of
transformer blocks, either from bottom or top of BERT.

We can observe that
(i) in most cases, top-down masking outperforms bottom-up masking
when initial sparsity and the number of masked layers are fixed.
Thus, it is reasonable to select all pretrained weights in lower layers,
since they capture general information helpful and transferable to various tasks
\citep{liu-etal-2019-linguistic,howard-ruder-2018-universal}.
(ii) For bottom-up masking, increasing the number of masked layers
gradually improves performance.
This observation illustrates dependencies between BERT layers
and the learning dynamics of masking: provided with
selected pretrained weights in lower layers, higher layers need to be given
flexibility to select pretrained weights accordingly to achieve good task performance.
(iii) In top-down masking, CoLA performance increases when masking a growing number of layers
while MRPC and RTE are not sensitive. Recall that CoLA tests linguistic acceptability
that typically requires both syntactic and semantic
information\footnote{ For example, to distinguish
	acceptable caused-motion constructions (e.g., ``the
	professor talked us into a stupor'')
	from inacceptable ones (e.g.,
	``water
	talked it into red''), both syntactic
	and semantic information need to be considered \cite{goldberg95construction}.
}.
All of BERT layers are involved in representing this
information, hence allowing more layers to change should improve performance.

\subsection{Comparing finetuning and masking}
\seclabel{maincomp}
We have investigated
two factors -- initial sparsity (\secref{initsparsity}) and
layer-wise behaviors (\secref{maskwhichlayers}) -- that are important in masking \ptls.
Here, we compare the performance and memory consumption of masking and
finetuning.

Based on observations in \secref{initsparsity} and \secref{maskwhichlayers},
we use 5\% initial sparsity when applying masking  to BERT, RoBERTa, and DistilBERT.
We mask the transformer blocks 2--11 in BERT/RoBERTa and 2--5 in DistilBERT. $\mW_{P}$ and $\mW_{T}$ are always masked.
Note that this global setup is surely suboptimal for some model-task combinations, but our goal
is to illustrate the effectiveness and the generalization ability of  masking.
Hence, conducting extensive hyperparameter search is unnecessary.

For AG and QNLI, we use batch size 128. For the other tasks we use batch size 32.
We search the optimal learning rate per task as described in \secref{datasetup},
and they are shown in Appendix \secref{learningratelist}.

\myparagraph{Performance comparison.} \tabref{devperf} reports performance
of masking and finetuning
on the dev set for the eleven NLP tasks.
We observe
that applying  masking to BERT/RoBERTa/DistilBERT yields
performance comparable to finetuning. We observe a performance drop\footnote{
	Similar observations were made:
	DistilBERT has a 10\% accuracy drop on RTE compared to BERT-base \citep{sanh2019distilbert};
	\citet{sajjad2020poor} report unstableness on MRPC and RTE when applying their model reduction strategies.}
on RoBERTa-RTE. RTE has the smallest dataset size (train: 2.5k; dev: 0.3k) among all tasks --
this may contribute to the imperfect results and large variances.

Our BERT-NER results are slightly worse than \citet{devlin-etal-2019-bert}. This
may be due to the fact that ``maximal document context'' is used by
\citet{devlin-etal-2019-bert} while we use sentence-level
context of 128 maximum
sequence length\footnote{Similar observations were made: \url{https://github.com/huggingface/transformers/issues/64}}.

Rows ``Single''  in \tabref{textcls} compare performance of
masking and finetuning BERT on the test set of SEM, TREC, AG, POS, and NER.
The same setup and hyperparameter searching as \tabref{devperf} are used,
the best hyperparameters are picked on the dev set.
Results from \citet{sun2019fine,palogiannidi-etal-2016-tweester} are
included as a reference.
\citet{sun2019fine} employ optimizations
like layer-wise learning rate,
producing slightly better performance than ours.
\citet{palogiannidi-etal-2016-tweester} is the best performing system on task
SEM \citep{nakov-etal-2016-semeval}.
Again, masking yields results comparable to finetuning.

\begin{figure}[t]
	\centering
	\subfloat[Number of parameters.]{
		\includegraphics[width=.45\linewidth]{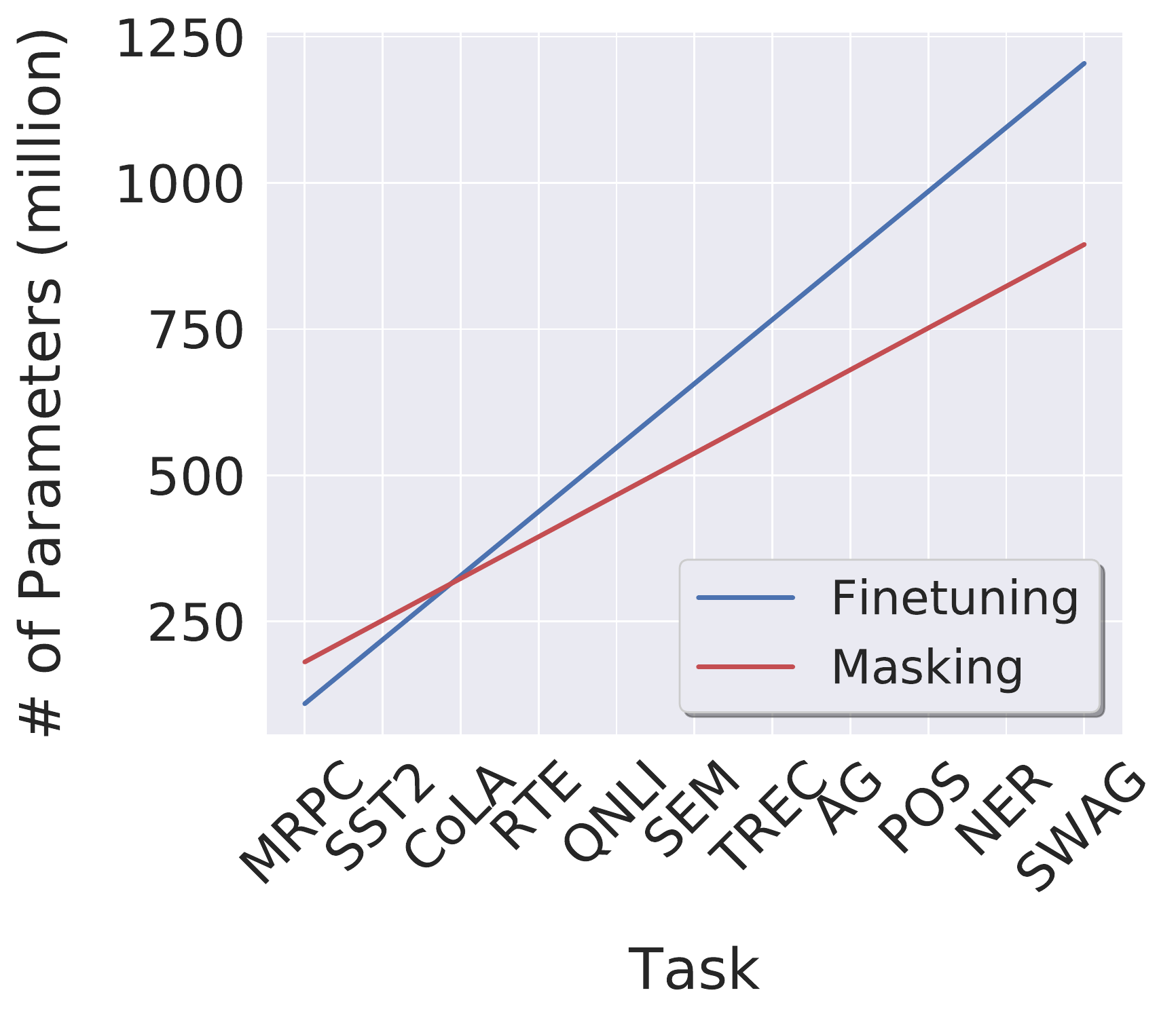}
	}
	\subfloat[Memory consumption.]{
		\includegraphics[width=.45\linewidth]{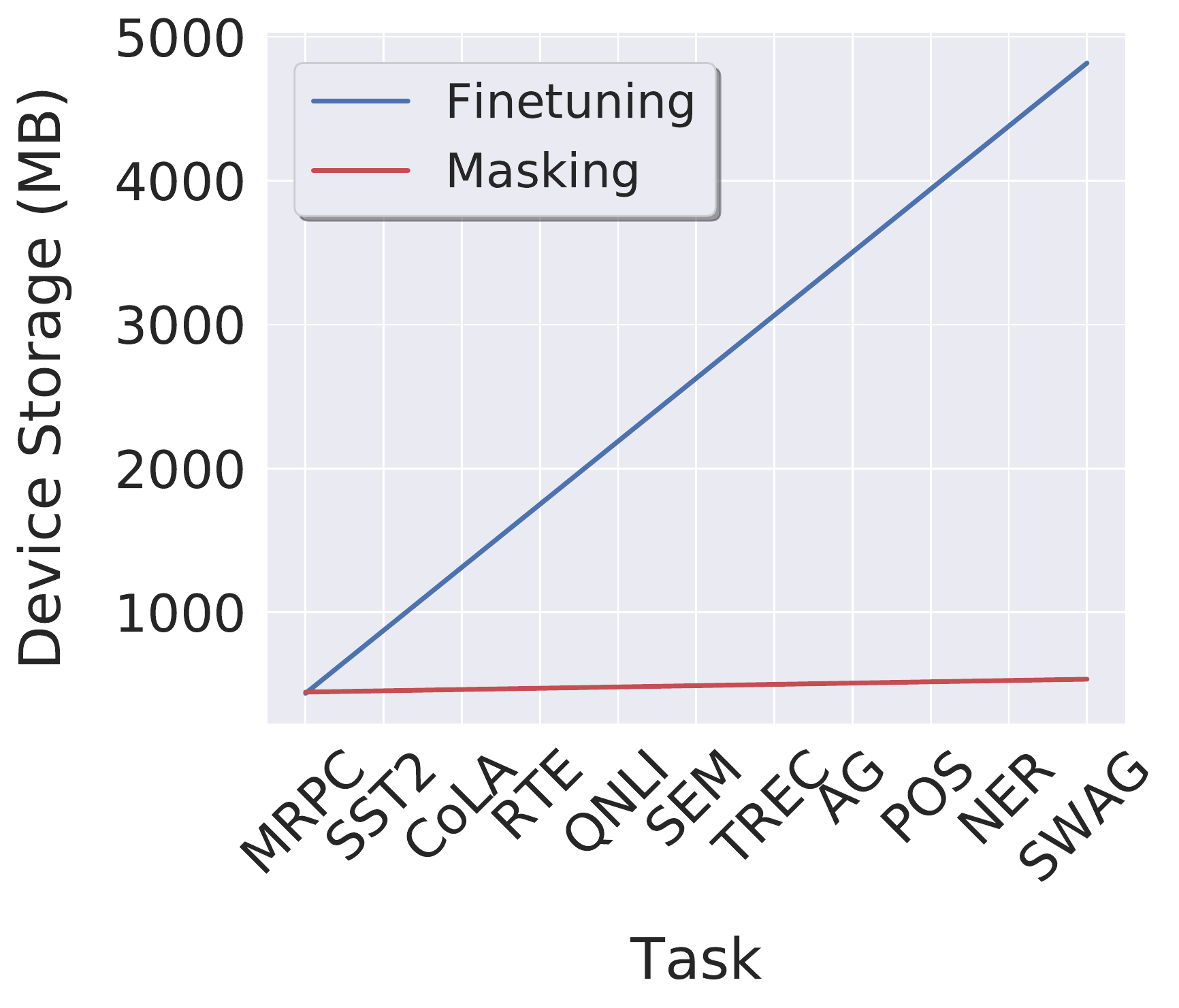}
	}
	\caption{
		The accumulated number of parameters and memory required by
		finetuning and masking to solve an increasing number of tasks.
	}
	\figlabel{memorycomp}
\end{figure}

\begin{table}[t]
	\centering
	\renewcommand{\arraystretch}{1.2}
	\resizebox{.48\textwidth}{!}{
		\begin{tabular}{cc|ccccc|c|}
			\cline{3-8}
			                                                &        & SEM   & TREC & AG   & POS  & NER  & Memory (MB) \\\hline
			\multicolumn{1}{|c|}{\multirow{2}{*}{Masking}}  & Single & 12.03 & 3.30 & 5.62 & 2.34 & 9.85 & 447         \\
			\multicolumn{1}{|c|}{}                          & Ensem. & 11.52 & 3.20 & 5.28 & 2.12 & 9.19 & 474         \\ \hline
			\multicolumn{1}{|c|}{\multirow{2}{*}{Finetun.}} & Single & 11.87 & 3.80 & 5.66 & 2.34 & 9.85 & 438         \\
			\multicolumn{1}{|c|}{}                          & Ensem. & 11.73 & 2.80 & 5.17 & 2.29 & 9.23 & 1752        \\ \hline
			\multicolumn{2}{|c|}{\citet{sun2019fine}}                      & n/a    & 2.80 & 5.25 & n/a & n/a & n/a               \\ \hline
			\multicolumn{2}{|c|}{\citet{palogiannidi-etal-2016-tweester}}  & 13.80  & n/a  & n/a  & n/a & n/a & n/a               \\ \hline
		\end{tabular}}
	\caption{
		Error rate (\%) on test set and model size comparison.
		Single: the averaged performance of four models with different random seeds.
		Ensem.: ensemble of the four models.
	}
	\tablabel{textcls}
\end{table}

\myparagraph{Memory comparison.}
Having shown that
task performance of masking and finetuning is comparable, we next
demonstrate one key strength of masking: memory efficiency. We take
BERT-base-uncased as our example.
\figref{memorycomp} shows the accumulated number of parameters in million and
memory in megabytes (MB) required when an increasing number of
downstream tasks need to be solved using finetuning and masking. Masking
requires a small overhead when solving a single task but is much more efficient
than finetuning when several tasks need to be inferred.
Masking saves a single copy of a \ptl containing 32-bit float parameters
for all the eleven tasks and a set of 1-bit binary masks for each task.
In contrast, finetuning saves every finetuned model
so the memory consumption grows linearly.

Masking  naturally allows light ensembles of models.
Rows ``Ensem.'' in \tabref{textcls} compare ensembled results and model size.
We consider the ensemble of predicted
(i) labels;
(ii) logits;
(iii) probabilities.
The best ensemble method is picked on  dev and then evaluated on test.
Masking  only consumes
474MB of memory -- much smaller than 1752MB required by finetuning -- and achieves
comparable performance.
Thus, masking is also much more memory-efficient than finetuning
in an ensemble setting.

\section{Discussion}
\subsection{Intrinsic evaluations}
\secref{experimentsec}
demonstrates that  masking is an efficient alternative to
finetuning. Now we analyze properties of the representations computed
by binary masked language models with intrinsic evaluation.

One intriguing property of finetuning, i.e.,
stacking a classifier layer on top of
a \ptl then update all parameters,
is that
a linear classifier layer suffices
to conduct reasonably accurate classification.
This observation implies that the configuration of data points, e.g.,
sentences with positive or negative sentiment in SST2, should be close to linearly
separable in the hidden space.
Like finetuning, masking also uses a linear classifier layer.
Hence, we hypothesize that upper layers in binary masked language models, even without explicit weight updating,
also create a hidden space in which data points are close to linearly separable.

\begin{figure}[t]
	\subfloat{
		\includegraphics[width=.5\linewidth]{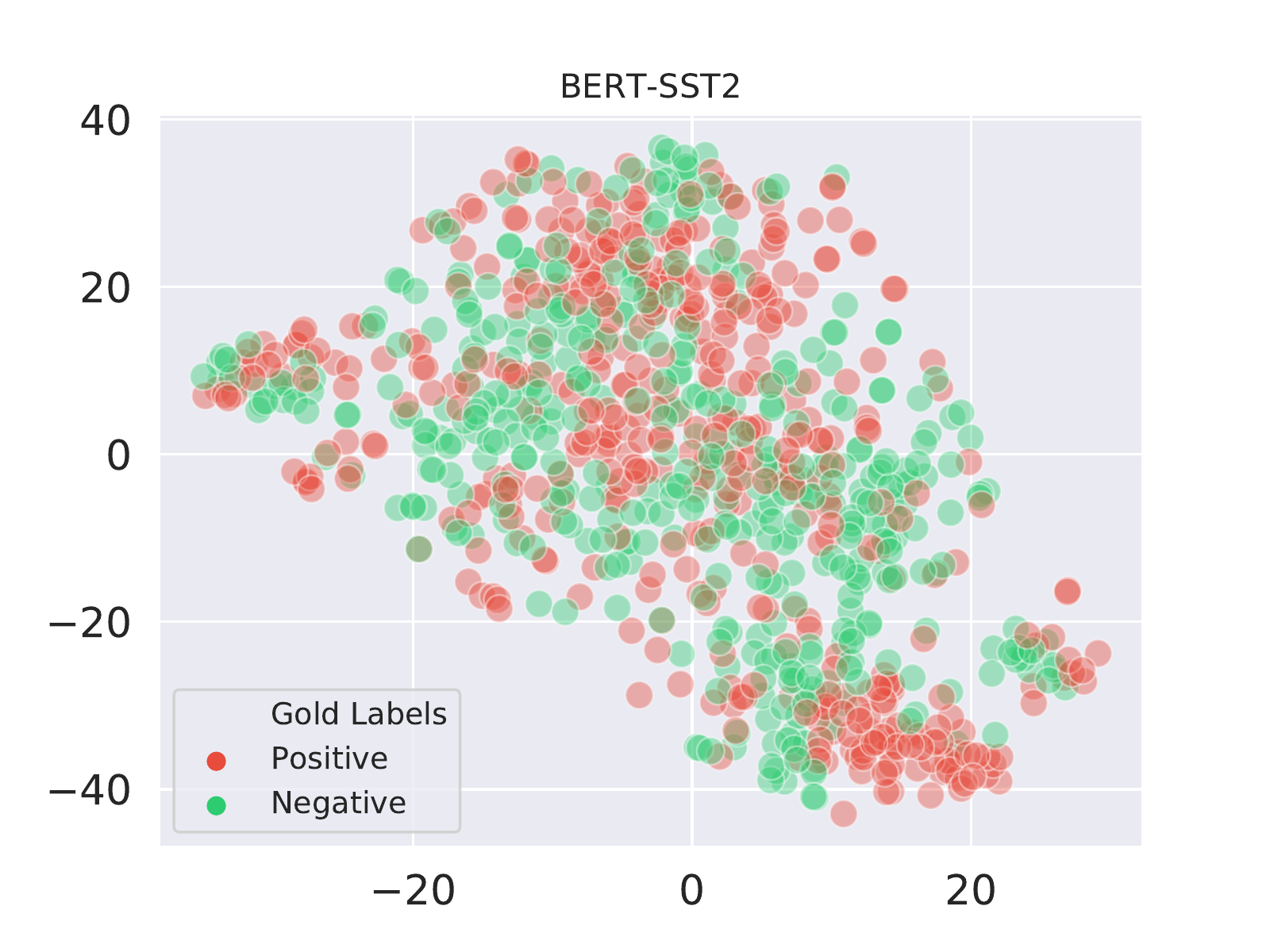}
		\includegraphics[width=.5\linewidth]{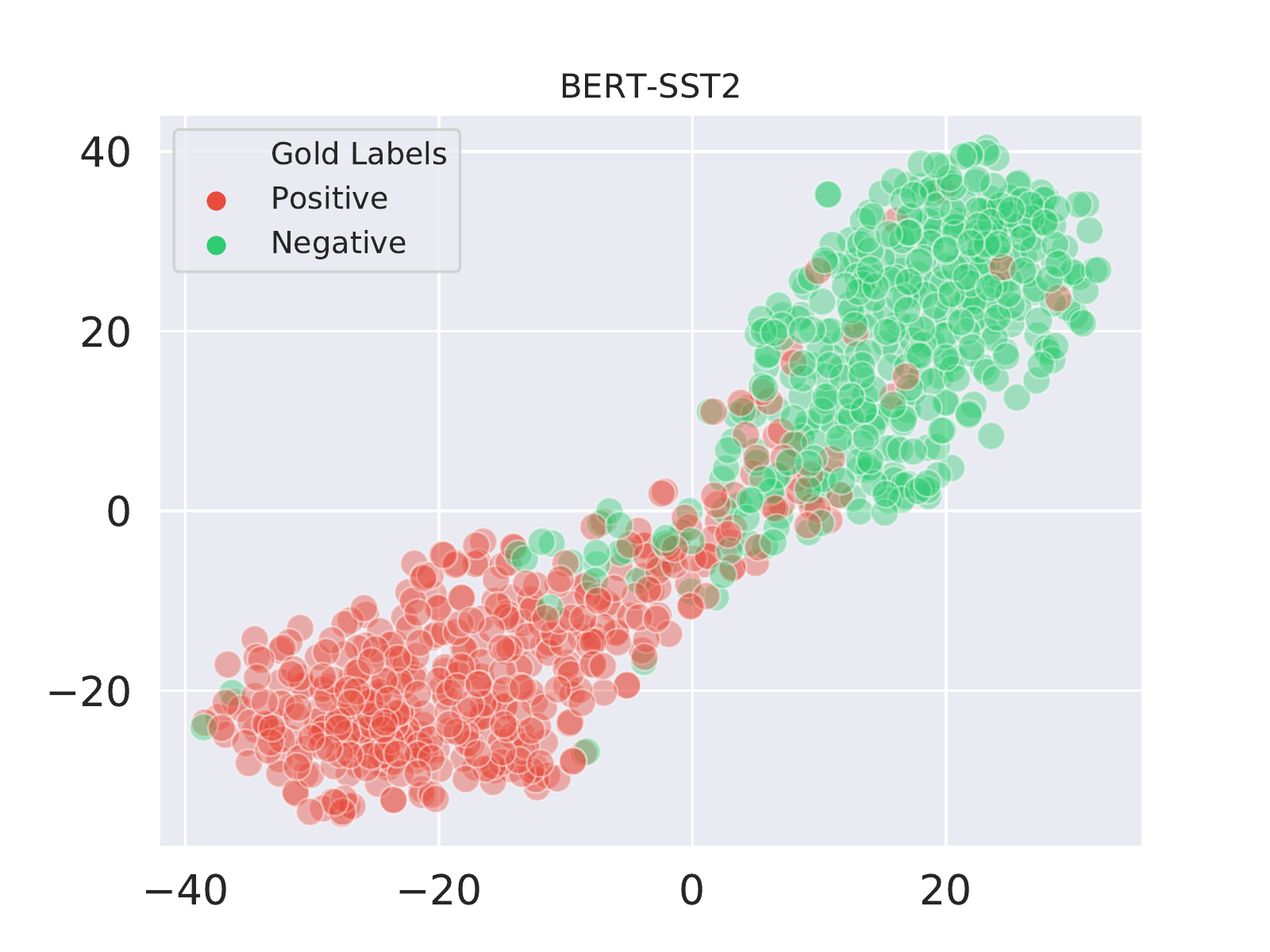}
	}
	\vfill
	\subfloat{
		\includegraphics[width=.5\linewidth]{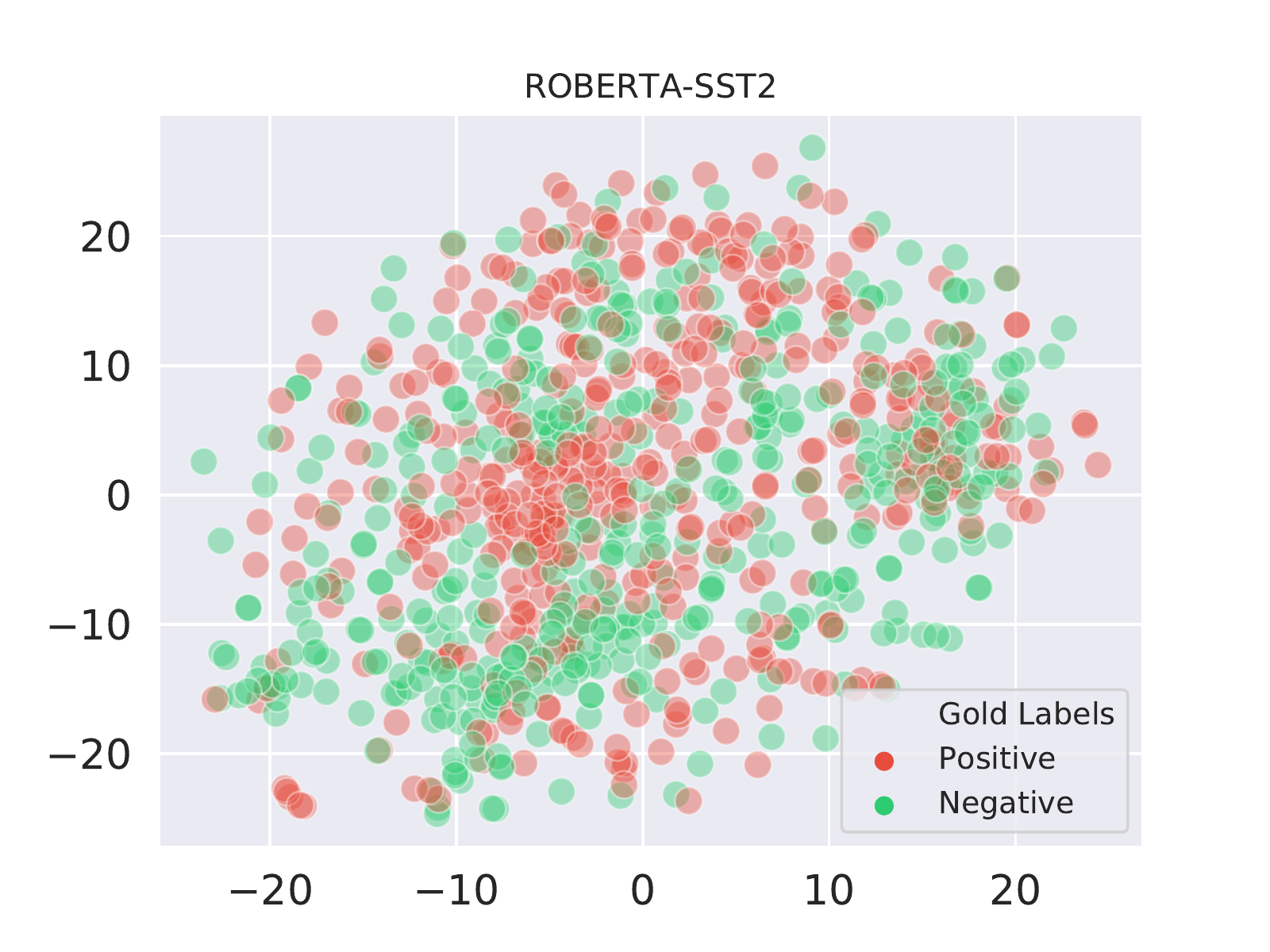}
		\includegraphics[width=.5\linewidth]{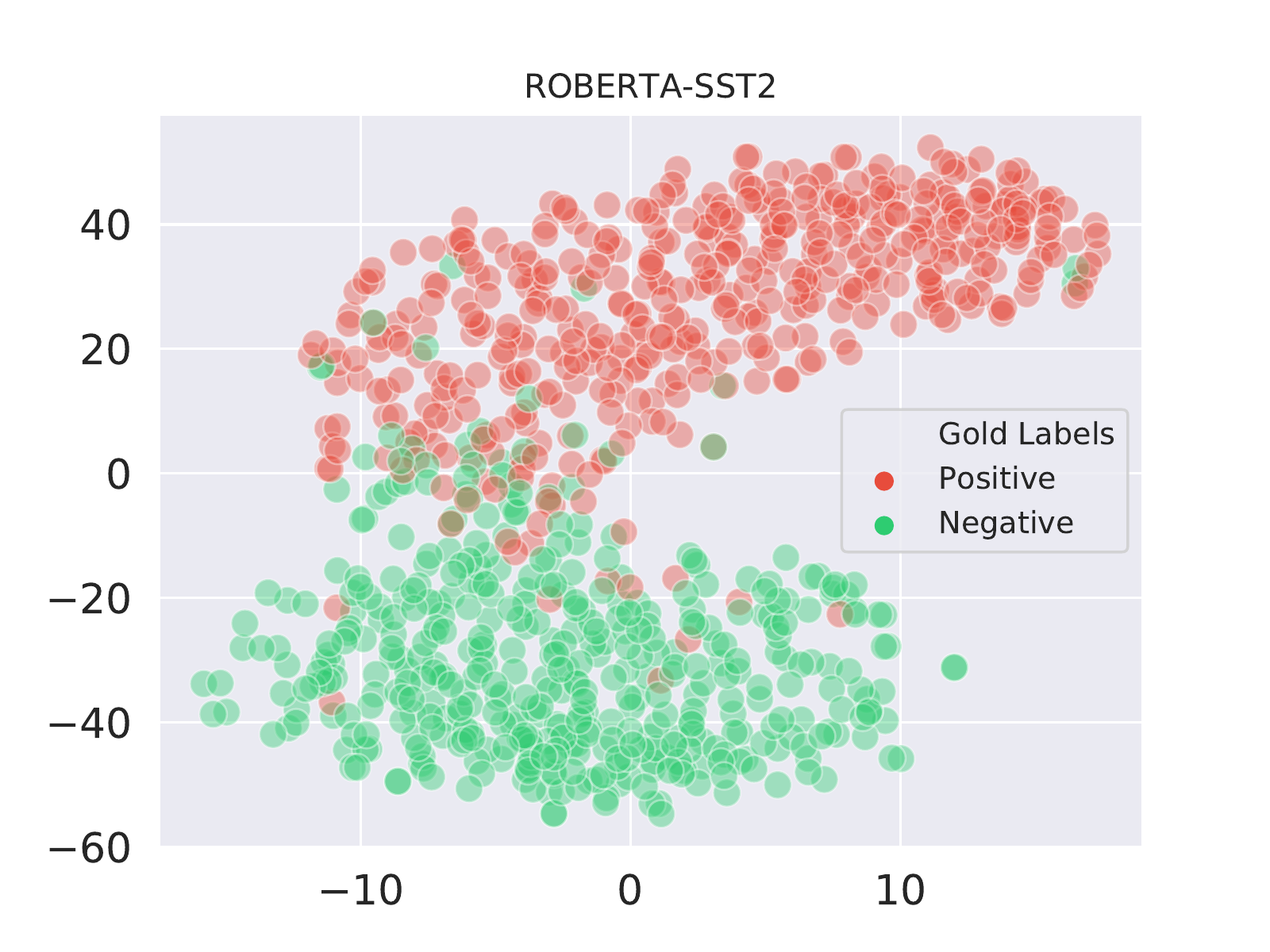}
	}
	\caption{
		t-SNE visualization of the representation of \texttt{[CLS]} computed
		by the topmost transformer block in pretrained (left), finetuned (top right), and
		masked (bottom right)
		BERT/RoBERTa.
		We
		use \texttt{scikit-learn} \citep{scikit-learn}
		and default t-SNE parameters.
	}
	\figlabel{tsne}
\end{figure}

\figref{tsne} uses t-SNE \citep{maaten2008visualizing} to visualize the
representation of \texttt{[CLS]} computed by the topmost transformer block in
pretrained, finetuned, and masked BERT/RoBERTa, using the dev set examples of SST2.
The pretrained models' representations (left) are clearly
not separable since the model needs to be adapted to downstream tasks.
The sentence representations computed by the finetuned (top right) and
the binary masked (bottom right) encoder are almost linearly separable and
consistent with the gold labels.
Thus, a linear
classifier is expected to yield reasonably good classification accuracy.
This intrinsic evaluation illustrates that binary masked models extract good
representations from the data for the downstream NLP task.

\begin{table}[t]
	\centering
	\renewcommand{\arraystretch}{1.2}
	\begin{minipage}{.2\textwidth}
		\small
		\resizebox{\textwidth}{!}{
			\begin{tabular}{c|cc|}
				\cline{2-3}
				                           & SST2 & SEM   \\ \hline
				\multicolumn{1}{|c|}{SST2} & 41.8 & -13.4 \\
				\multicolumn{1}{|c|}{SEM}  & 20.0 & 11.5  \\ \hline
			\end{tabular}
		}
		\caption*{(a) Masking}
	\end{minipage}
	\quad
	\begin{minipage}{.2\textwidth}
		\small
		\resizebox{\textwidth}{!}{
			\begin{tabular}{c|cc|}
				\cline{2-3}
				                           & SST2 & SEM   \\ \hline
				\multicolumn{1}{|c|}{SST2} & 41.8 & -10.1 \\
				\multicolumn{1}{|c|}{SEM}  & 18.9 & 12.2  \\ \hline
			\end{tabular}
		}
		\caption*{(b) Finetuning}
	\end{minipage}
	\caption{
		Generalization on dev (\%) of binary masked and finetuned BERT.
		Row: training dataset; Column: evaluating dataset.
		Numbers are improvements against
		the majority-vote baseline: 50.9 for SST2 and 74.4 for SEM. Results are averaged
		across four random seeds.}
	\tablabel{crossdata}
\end{table}

\subsection{Properties of the binary masked models}
\myparagraph{Do binary masked models generalize?}
\figref{tsne} shows that a binary masked language
model produces proper representations for the
classifier layer and hence
performs as well as a finetuned model. Here, we are
interested in verifying that the binary masked model
does indeed solve downstream tasks
by learning meaningful representations --
instead of exploiting spurious correlations
that generalize poorly \cite{niven-kao-2019-probing,mccoy-etal-2019-right}.
To this end, we test if the binary masked mode is
generalizable to other datasets of the same type of downstream task.
We use
the two sentiment classification datasets:
SST2 and SEM. We simply evaluate the model masked or finetuned on SST2 against the
dev set of SEM and vice versa.
\tabref{crossdata} reports the results against the majority-vote baseline.
The finetuned and binary masked models of SEM generalize well on SST2,
showing $\approx$ 20\% improvement against the majority-vote baseline.

On the other hand, we observe that the knowledge learned on SST2 does not
generalize to SEM, for both finetuning and masking.
We hypothesize that this is because the Twitter domain (SEM) is much
more specific than movie reviews (SST2). For example, some Emojis or
symbols like ``:)'' reflecting strong sentiment do not occur in SST2,
resulting in unsuccessful generalization.
To test our hypothesis, we take another movie review dataset IMDB
\citep{maas-etal-2011-learning}, and directly apply the SST2-finetuned- and
SST2-binary-masked- models on it.
Masking and finetuning achieve accuracy 84.79\% and 85.25\%, which
are comparable and both outperform the baseline 50\%, demonstrating
successful knowledge transfer.

Thus, finetuning and masking yield models with similar generalization
ability. The binary masked models indeed create representations that contain
valid information for downstream tasks.

\begin{figure}[t]
	\centering

	\subfloat{
		\includegraphics[width=.5\columnwidth]{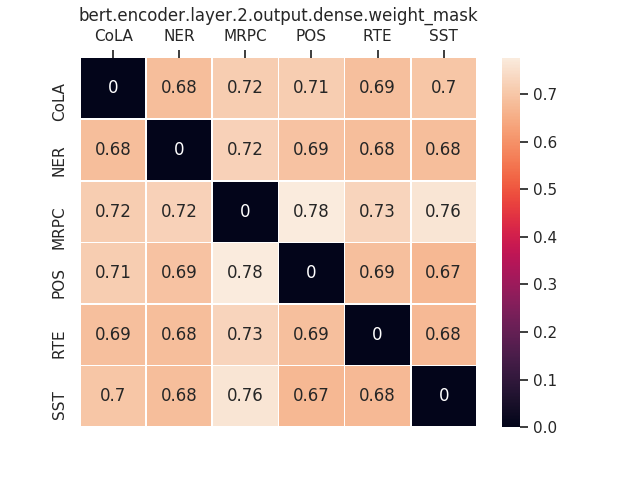}
	}
	\subfloat{
		\includegraphics[width=.5\columnwidth]{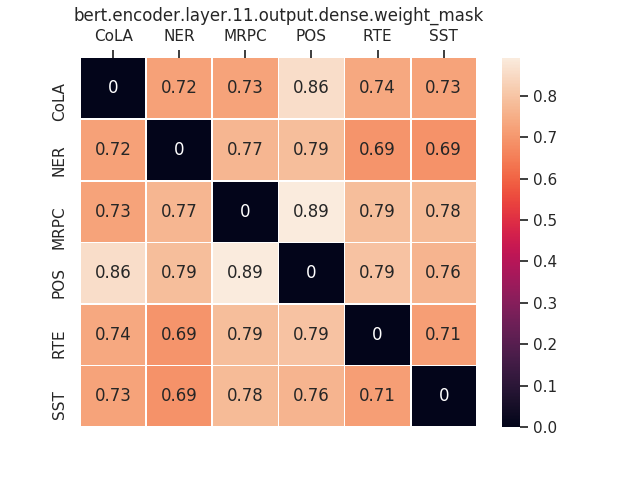}
	}
	\caption{
		Scores $s$ of two sets of masks, trained with two different tasks, of layer $\mW_O$
		in transformer blocks 2 (left) and 11 (right) in BERT.
		A large $s$ means that the two  masks are  dissimilar.}
	\figlabel{maskrelation}
\end{figure}

\myparagraph{Analyzing masks.}
We study the dissimilarity
between masks learned by
different
BERT layers and downstream tasks.
For
the initial and trained binary masks $\mMb^{t, init}$ and $\mMb^{t, trained}$
of a layer trained on task $t \in \{t1, t2\}$. We compute:
\begin{align*}
	\textstyle
	s = \frac{\norm{\mMb^{t1, trained} - \mMb^{t2, trained}}_1}{
	\norm{\mMb^{t1, trained} - \mMb^{t1, init}}_1 + \norm{\mMb^{t2, trained} - \mMb^{t2, init}}_1} \,,
\end{align*}
where $\norm{\mW}_1 \!=\! \sum_{i=1}^{m} \sum_{j=1}^{n} \abs{w_{i, j}}$. %
Note that for the same random seed,
$\mMb^{t1, init}$ and $\mMb^{t2, init}$ are the same. The
dissimilarity
$s$
measures
the difference between two masks as a fraction of all changes
brought about by training.
\figref{maskrelation} shows that,
after training, the dissimilarities of  masks of higher BERT layers
are larger than those of lower BERT layers.
Similar observations are made
for finetuning: top layer weights in finetuned BERT are
more task-specific \citep{kovaleva2019revealing}. The figure
also shows that
the learned masks for downstream tasks tend to be dissimilar to each other, even for
similar tasks.
For a given task, there exist different sets of masks (initialized with different random seeds) yielding similar performance.
This observation is similar to the results of evaluating
the lottery ticket hypothesis on BERT \citep{prasanna2020bert,chen2020lottery}:
a number of subnetworks exist in BERT achieving similar task performance.

\subsection{Loss landscape}
Training complex neural networks can be viewed as searching
for good minima in the
highly non-convex landscape defined by the loss function
\citep{li2018visualizing}.
Good minima are typically depicted as points at the bottom
of different locally convex valleys \citep{keskar2016large,draxler2018essentially}, achieving
similar performance. In this section,
we study the
relationship between the
two
minima obtained by masking and finetuning.

Recent work analyzing the loss landscape suggests
that the local minima
in the loss landscape reached by
standard training algorithms
can be connected by a simple path~\citep{garipov2018loss,gotmare2018closer},
e.g., a B\'{e}zier curve,
with low task loss (or high task accuracy) along the path.
We are interested in testing if the two minima found by finetuning and masking
can be easily connected in the loss landscape.
To start with, we verify
the task performance of an interpolated model $\mW(\gamma)$ on the line segment between
a finetuned model $\mW_0$ and a binary masked model $\mW_1$:
\begin{align*}
	\mW(\gamma) = \mW_0 + \gamma(\mW_1 - \mW_0), 0\leq \gamma \leq 1 \,.
\end{align*}

We conduct experiments on MRPC and SST2 with the best-performing BERT and RoBERTa models
obtained in \tabref{devperf} (same seed and training epochs);
\figref{linear_mode_connectivity} (top) shows the results of mode connectivity, i.e., the evolution of the task accuracy along a line connecting  the two candidate minima.

Surprisingly, the interpolated models on the line segment connecting a finetuned and
a binary masked model form a high accuracy path,
indicating the extremely well-connected loss landscape. Thus, masking
finds minima on the same connected low-loss manifold as finetuning,
confirming the effectiveness of our method.
Also, we show in~\figref{linear_mode_connectivity} (bottom)
for the line segment between the pretrained BERT and a
finetuned/masked BERT, that
mode connectivity is not solely due to an overparameterized \ptl.
Bézier curves experiments show similar results, cf. Appendix
\secref{appendix_mode_connectivity}.

\begin{figure}[t]
	\subfloat{
		\includegraphics[width=.45\linewidth]{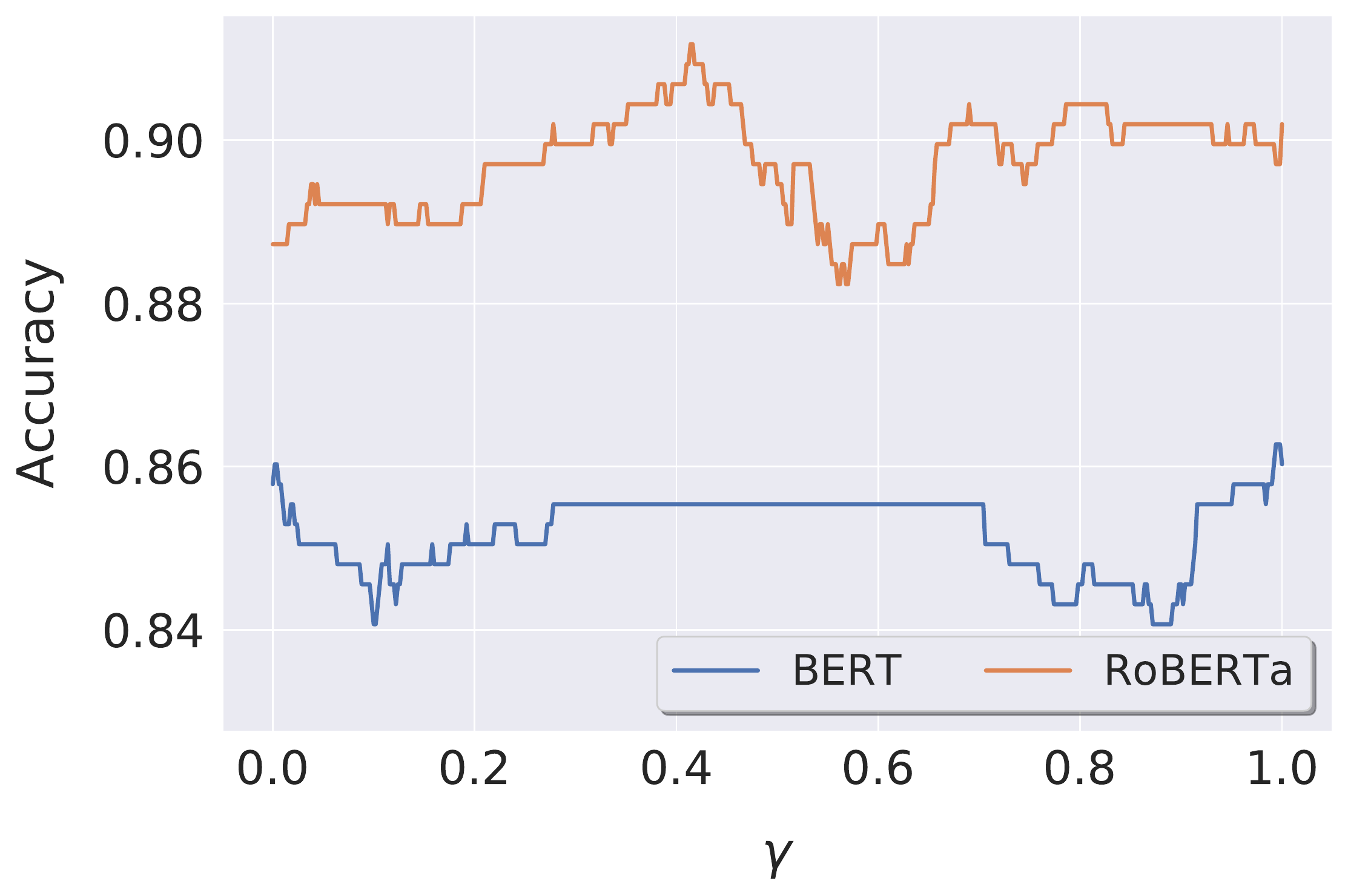}
	}
	\subfloat{
		\includegraphics[width=.45\linewidth]{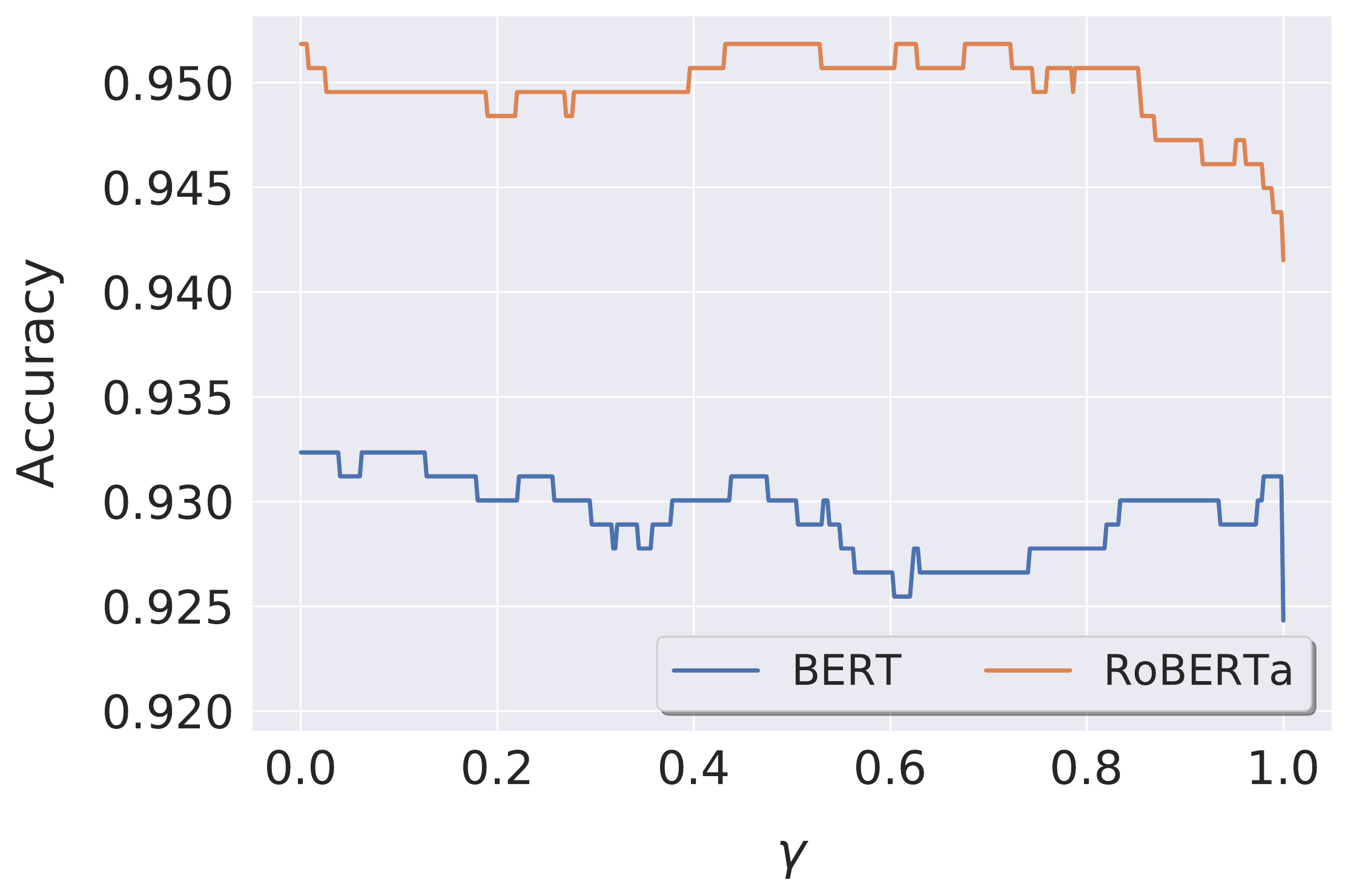}
	}
	\\
	\subfloat{
		\includegraphics[width=.45\linewidth]{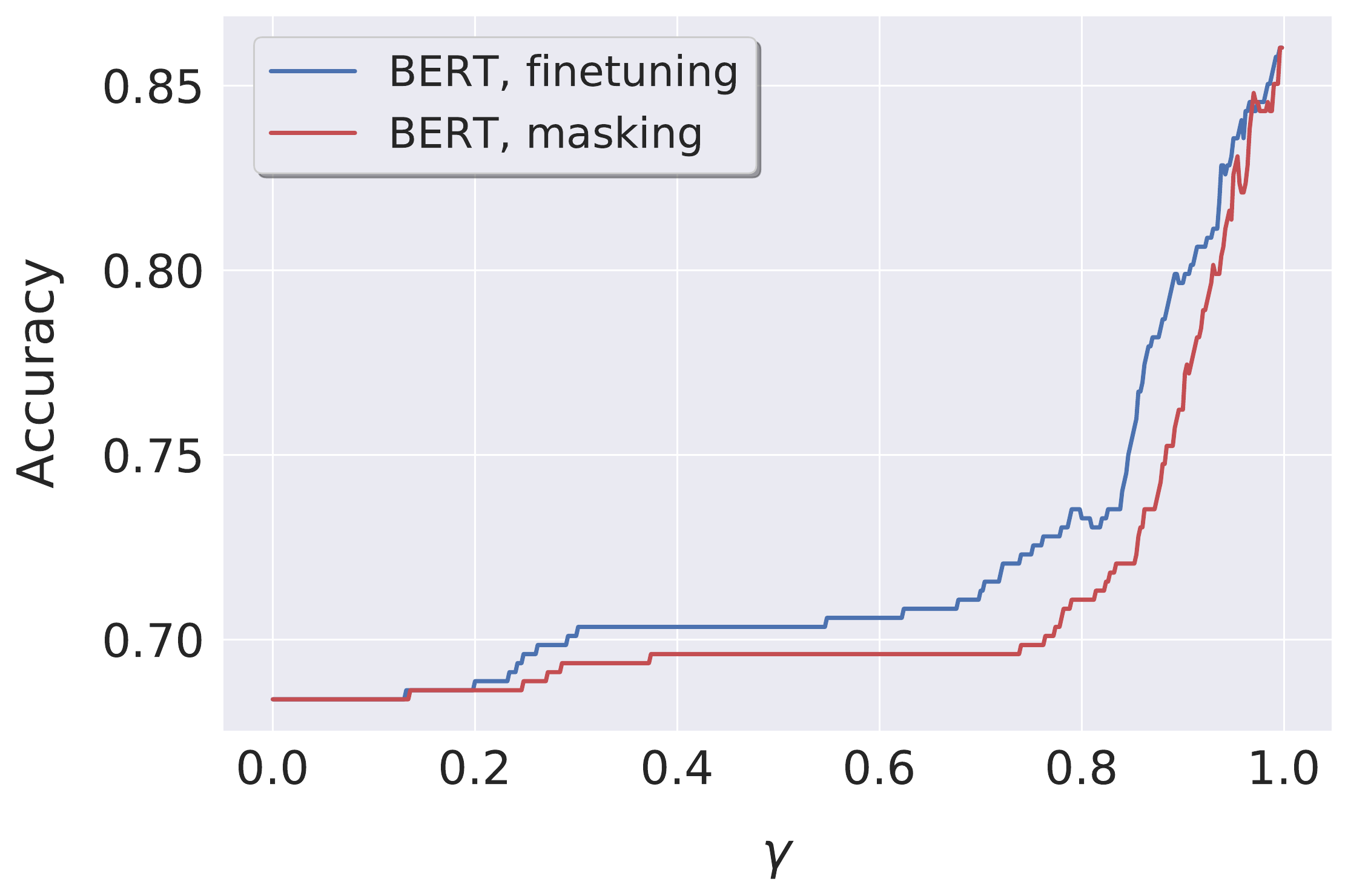}
	}
	\subfloat{
		\includegraphics[width=.45\linewidth]{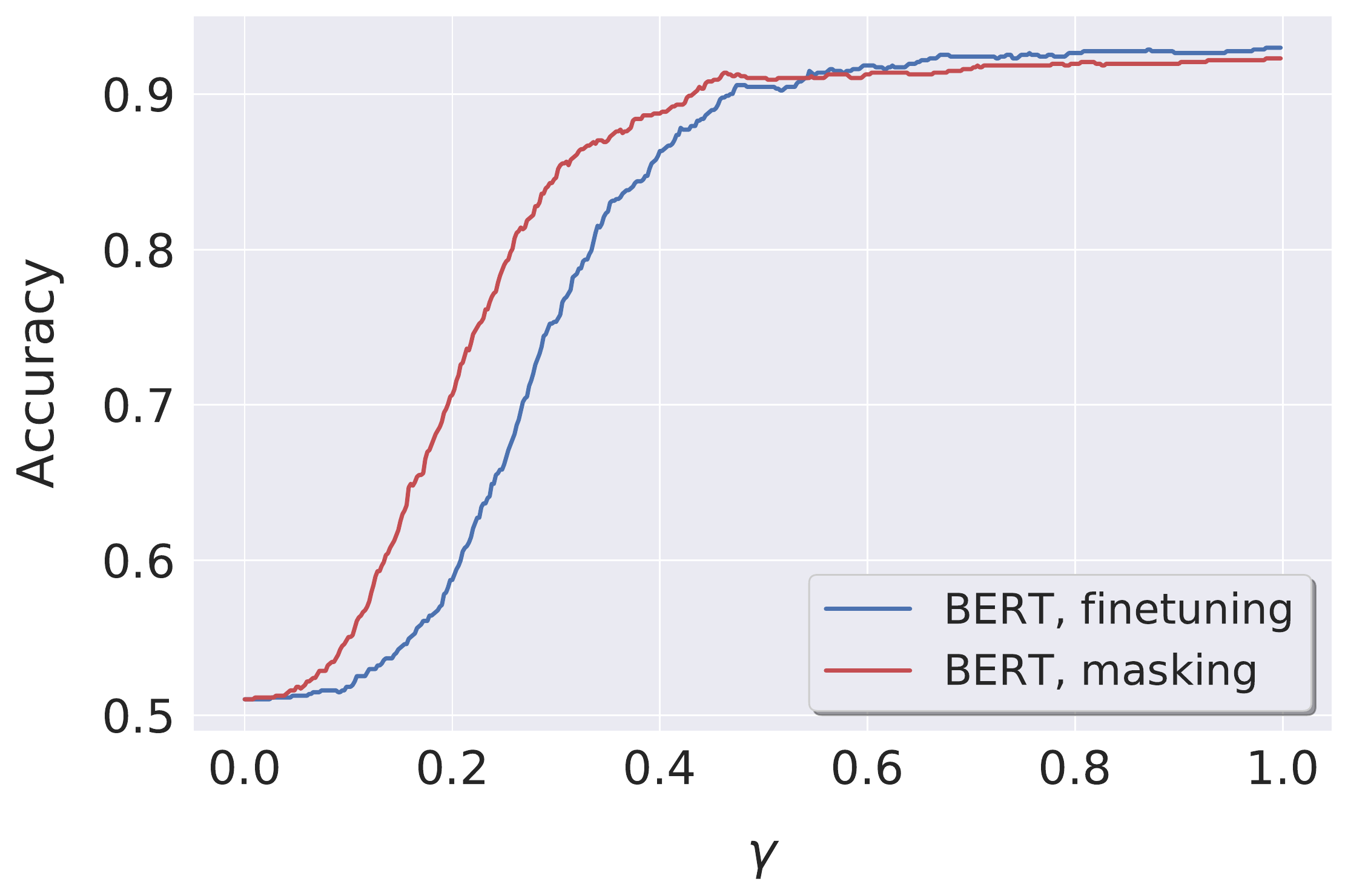}
	}
	\caption{
		Mode connectivity results on MRPC (left) and SST2 (right).
		Top images: dev set accuracy of an interpolated model between
		the two minima found by finetuning ($\gamma\!=\!0$) and masking ($\gamma\!=\!1$).
		Bottom images: accuracy of an interpolated model between pretrained ($\gamma\!=\!0$) and finetuned/masked ($\gamma\!=\!1$) BERT.
	}

	\figlabel{linear_mode_connectivity}
\end{figure}

\section{Conclusion}
We have presented masking, an efficient alternative to finetuning for utilizing \ptls like
BERT/RoBERTa/DistilBERT.
Instead of updating the pretrained parameters,
we only train one set of binary masks per task to select critical parameters.
Extensive experiments show that masking
yields performance comparable to finetuning on a series of NLP tasks.
Leaving the pretrained parameters unchanged, masking is much more
memory efficient when several tasks need to be solved.
Intrinsic evaluations show
that binary masked models extract valid and generalizable representations for downstream tasks.
Moreover, we demonstrate that the minima obtained by finetuning and masking can be
easily connected by a line segment, confirming the effectiveness of applying
masking to \ptls. Our code is available at: \url{https://github.com/ptlmasking/maskbert}.

Future work may explore the possibility of applying masking to the pretrained
multilingual encoders like mBERT \citep{devlin-etal-2019-bert} and
XLM \citep{conneau2019cross}. Also, the binary masks learned by our
method have low sparsity such that inference speed is not improved.
Developing methods improving both memory and inference efficiency
without sacrificing task performance can open the possibility of
widely deploying the powerful \ptls to more NLP applications.

\section*{Acknowledgments}
We thank
the anonymous reviewers for the insightful comments and suggestions.
This work was funded by the European Research Council (ERC \#740516),
SNSF grant 200021\_175796, as well as a Google Focused Research Award.

\bibliography{emnlp2020}
\bibliographystyle{acl_natbib}

\clearpage
\appendix

\section{ Reproducibility Checklist}
\seclabel{checklist}
\subsection{Computing infrastructure} All experiments are conducted on following
GPU models: Tesla V100, GeForce GTX 1080 Ti, and GeForce GTX 1080.
We use per-GPU batch size 32. Thus, experiments comparing masking and finetuning
on QNLI and AG take 4 GPUs and all the other tasks use a single GPU.

\subsection{Number of parameters}
In \secref{maincomp} we thoroughly compare the
number of parameters and memory consumption of finetuning and masking.
Numerical values are in \tabref{tab:memorycomp}.

\subsection{Validation performance}
The dev set performance of \tabref{textcls} is covered in \tabref{devperf}.
We report Matthew's correlation coefficient (MCC) for CoLA,
micro-F1 for NER, and accuracy for the other tasks. We use the evaluation
functions in \texttt{scikit-learn} \citep{scikit-learn} and \texttt{seqeval}
(\url{https://github.com/chakki-works/seqeval}).

\subsection{Hyperparameter search}
\seclabel{learningratelist}
The only hyperparameter we searched is learning
rate, for both masking and finetuning, according to the setup discussion in
\secref{datasetup}. The optimal values are in \tabref{baseline_hyperparameter_lr}.

\begin{table*}[!h]
	\centering
	\renewcommand{\arraystretch}{1.2}
	\resizebox{.9\textwidth}{!}{%
		\begin{tabular}{cc|ccccccccccc|}
			\cline{3-13}
			                                                  &            & MRPC & SST2 & CoLA & RTE  & QNLI & POS  & NER  & SWAG & SEM  & TREC & AG   \\ \hline
			\multicolumn{1}{|c|}{\multirow{2}{*}{BERT}}       & Finetuning & 5e-5 & 1e-5 & 3e-5 & 5e-5 & 3e-5 & 3e-5 & 3e-5 & 7e-5 & 1e-5 & 3e-5 & 3e-5 \\
			\multicolumn{1}{|c|}{}                            & Masking    & 1e-3 & 5e-4 & 9e-4 & 1e-3 & 7e-4 & 5e-4 & 7e-4 & 1e-4 & 7e-5 & 1e-4 & 5e-4 \\ \hline
			\multicolumn{1}{|c|}{\multirow{2}{*}{RoBERTa}}    & Finetuning & 3e-5 & 1e-5 & 1e-5 & 7e-6 & 1e-5 & 9e-6 & 3e-5 & 1e-5 & 7e-6 & 9e-6 & 3e-5 \\
			\multicolumn{1}{|c|}{}                            & Masking    & 3e-4 & 9e-5 & 3e-4 & 3e-4 & 1e-4 & 3e-4 & 3e-4 & 1e-4 & 3e-4 & 5e-4 & 5e-4 \\\hline
			\multicolumn{1}{|c|}{\multirow{2}{*}{DistilBERT}} & Finetuning & 3e-5 & 7e-5 & 3e-5 & 3e-5 & 3e-5 & 3e-5 & 1e-5 & 7e-6 & 1e-5 & 3e-5 & 3e-5 \\
			\multicolumn{1}{|c|}{}                            & Masking    & 9e-4 & 7e-4 & 9e-4 & 9e-4 & 1e-3 & 7e-4 & 7e-4 & 3e-4 & 3e-4 & 9e-4 & 1e-3 \\\hline
		\end{tabular}}
	\caption{
		The optimal learning rate on different tasks for BERT/RoBERTa/DistilBERT.
		We perform finetuning/masking on all tasks for 10 epochs with early stopping of 2 epochs.
	}
	\tablabel{baseline_hyperparameter_lr}
\end{table*}

\subsection{Datasets} For GLUE tasks, we use the official datasets from
the benchmark \url{https://gluebenchmark.com/}.
For TREC and AG, we download the datasets developed by \citet{NIPS2015_5782}, which are available at
\href{https://drive.google.com/drive/u/0/folders/0Bz8a_Dbh9Qhbfll6bVpmNUtUcFdjYmF2SEpmZUZUcVNiMUw1TWN6RDV3a0JHT3kxLVhVR2M}{here.}
Note that this link is provided by \citet{NIPS2015_5782} and also used by \citet{sun2019fine}.
For SEM, we obtain the dataset from the official SemEval website: \url{http://alt.qcri.org/semeval2016/task4/}.
For NER, we use the official dataset: \url{https://www.clips.uantwerpen.be/conll2003/ner/}.
We obtain our POS dataset from the linguistic data consortium (LDC).
We use the official dataset of SWAG \citep{zellers-etal-2018-swag}: \url{https://github.com/rowanz/swagaf/tree/master/data}.

For POS, sections 0-18 of WSJ are train, sections 19-21 are dev, and sections 22-24 are test \citep{collins-2002-discriminative}.
We use the official train/dev/test splits of all the other datasets.

To preprocess the datasets, we use the tokenizers provided by the \texttt{Transformers} package \citep{Wolf2019HuggingFacesTS}
to convert the raw dataset to the formats required by BERT/RoBERTa/DistilBERT.
Since wordpiece tokenization is used, there is no out-of-vocabulary words.

Since we use a maximum sequence length of 128, our preprocessing steps exclude some word-tag annotations in POS and NER.
For POS, after wordpiece tokenization, we see 1 sentence in dev and 2 sentences in test have more than 126
(the \texttt{[CLS]} and \texttt{[SEP]} need to be considered) wordpieces.
As a result, we exclude 5 annotated words in dev and 87 annotated words in test.
Similarly, for NER (which is also formulated as a tagging task following \citet{devlin-etal-2019-bert}),
we see 3 sentences in dev and 1 sentence in test have more than 126
wordpieces. As a result, we exclude 27 annotated words in dev and 8 annotated words in test.

The number of examples in dev and test per task is shown in following \tabref{completedatastats}.

\begin{table}[t]
	\centering
	\renewcommand{\arraystretch}{1.2}
	\begin{tabular}{c|rr|}
		\cline{2-3}
		                           & Dev     & Test    \\ \hline
		\multicolumn{1}{|c|}{MRPC} & 408     & n/a     \\
		\multicolumn{1}{|c|}{SST2} & 872     & n/a     \\
		\multicolumn{1}{|c|}{CoLA} & 1,042   & n/a     \\
		\multicolumn{1}{|c|}{RTE}  & 277     & n/a     \\
		\multicolumn{1}{|c|}{QNLI} & 5,732   & n/a     \\
		\multicolumn{1}{|c|}{SEM}  & 1,325   & 10,551  \\
		\multicolumn{1}{|c|}{TREC} & 548     & 500     \\
		\multicolumn{1}{|c|}{AG}   & 24,000  & 7,600   \\
		\multicolumn{1}{|c|}{POS}  & 135,105 & 133,082 \\
		\multicolumn{1}{|c|}{NER}  & 51,341  & 46,425  \\
		\multicolumn{1}{|c|}{SWAG} & 20,006  & n/a     \\ \hline
	\end{tabular}
	\caption{Number of examples in dev and test per task. For POS and NER, we report
		the number of words.}
	\tablabel{completedatastats}
\end{table}

\section{More on Mode Connectivity} \seclabel{appendix_mode_connectivity}
Following the mode connectivity framework proposed in~\citet{garipov2018loss},
we parameterize the path joining two minima using a Bézier curve.
Let $\ww_0$ and $\ww_{n+1}$ be the parameters of the models trained from finetuning and masking.
Then, an $n$-bend Bézier curve connecting $\ww_0$ and $\ww_{n + 1}$,
with $n$ trainable intermediate models $\mtheta = \{ \ww_1, \ldots, \ww_n \}$,
can be represented by $\phi_\mtheta (t)$,
such that $\phi_\mtheta (0) = \ww_0$ and $\phi_\mtheta (1) = \ww_{n+1}$,
and
\begin{align*}
	\small
	\phi_\mtheta (t) = \sum_{i=0}^{n + 1} {n + 1 \choose i} (1 - t)^{n + 1 - i} t^i \ww_i \,.
\end{align*}

We train a $3$-bend Bézier curve by minimizing the loss
$\E_{ t \sim U[0, 1] }{} \cL \left( \phi_\mtheta (t) \right) $,
where $U[0, 1]$ is the uniform distribution in the interval $[0, 1]$.
Monte Carlo method is used to estimate the gradient of this expectation-based function
and gradient-based optimization is used for the minimization.
The results are illustrated in \figref{complex_mode_connectivity}.
Masking implicitly performs gradient descent, analogy to the weights update achieved by finetuning;
the observations complement our arguments in the main text.

\begin{figure*}
	\centering
	\subfloat[BERT]{
		\includegraphics[width=.5\linewidth]{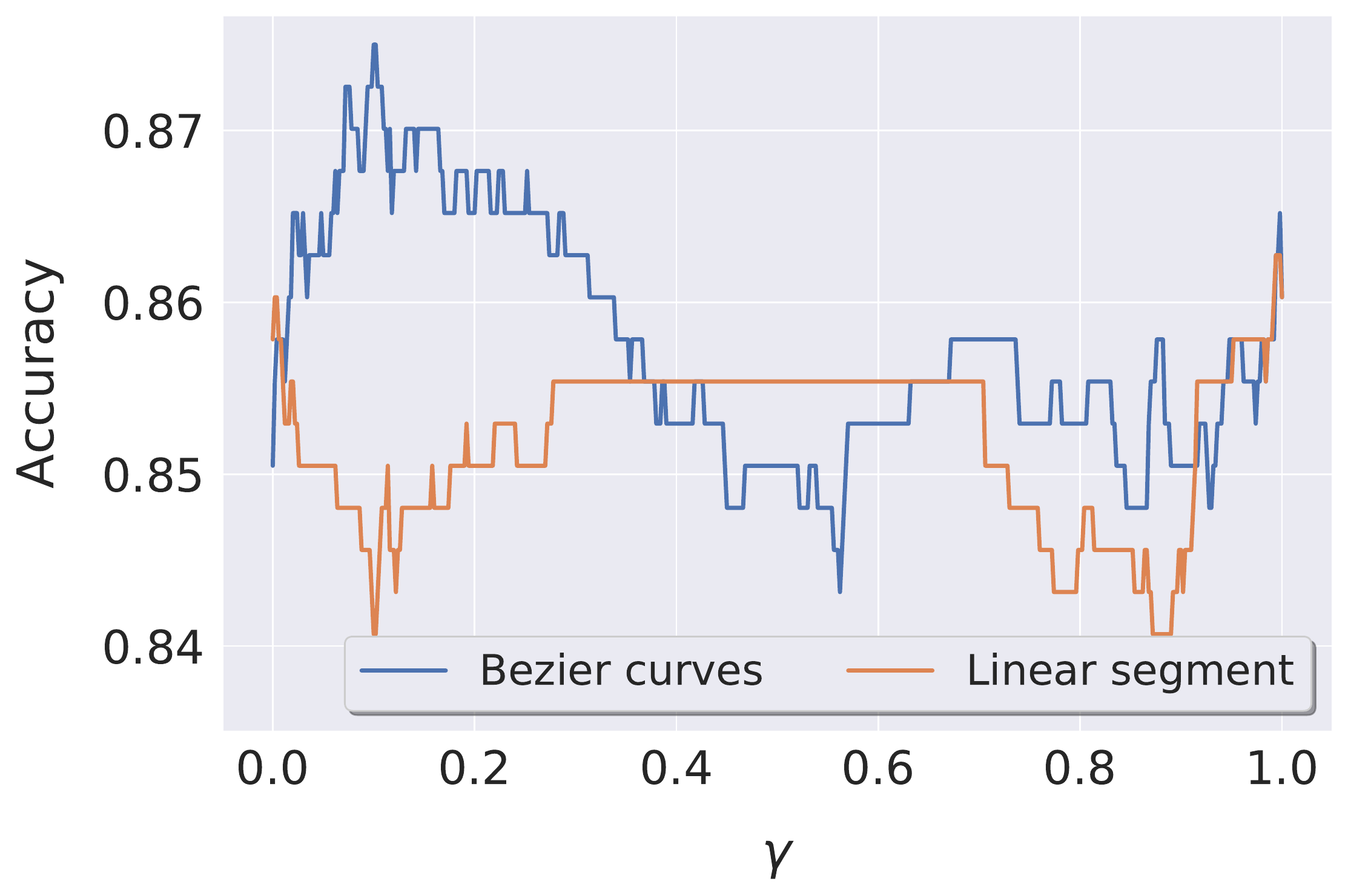}
	}
	\subfloat[RoBERTa]{
		\includegraphics[width=.5\linewidth]{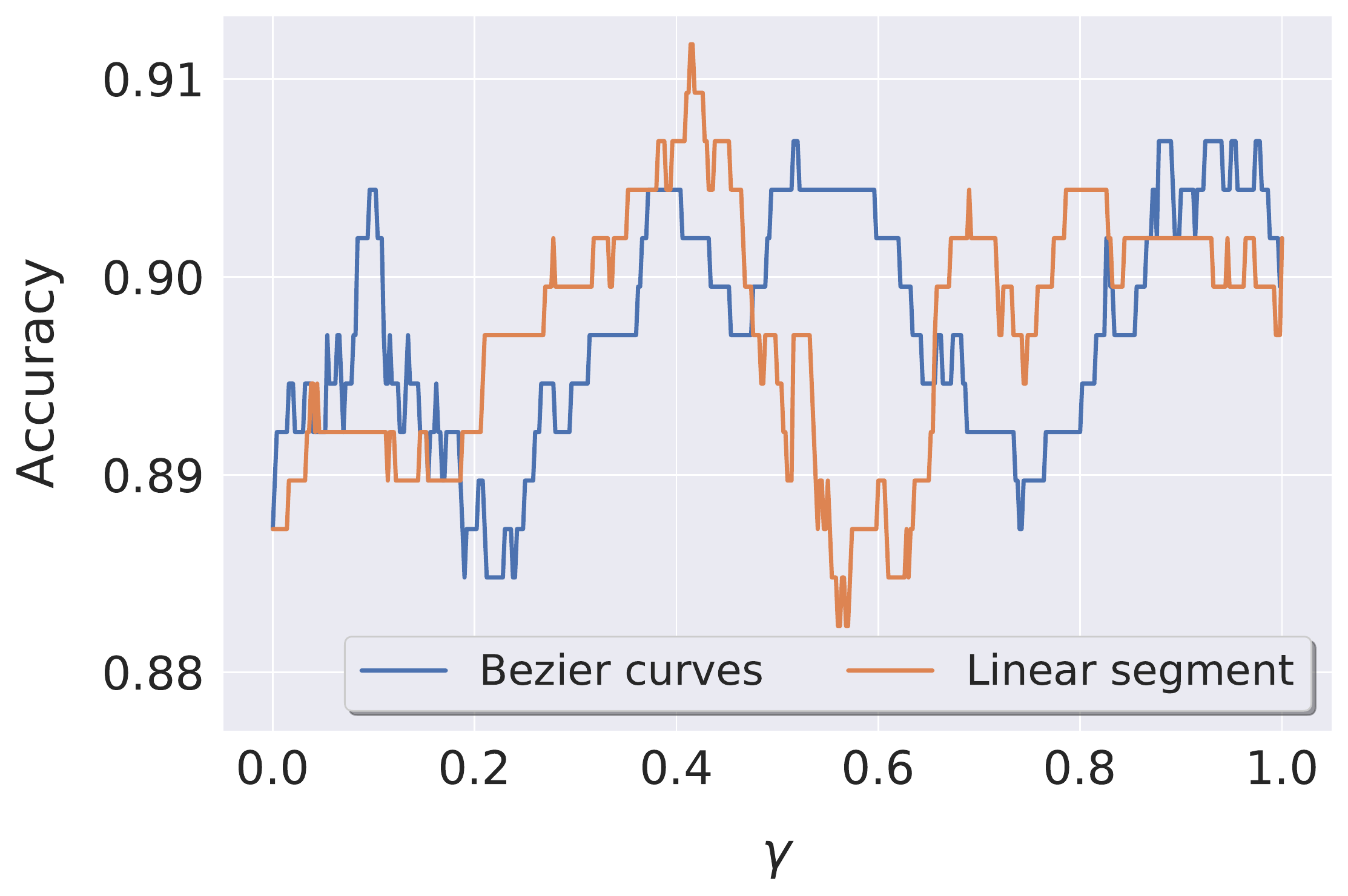}
	}
	\caption{
		The accuracy on MRPC dev set,
		as a function of the point on the curves $\phi_{\theta}(\gamma)$,
		connecting the two minima found by finetuning (left, $\gamma \!=\! 0$) and masking (right, $\gamma \!=\! 1$).
	}
	\figlabel{complex_mode_connectivity}
\end{figure*}

\section{More Empirical Results}
\textbf{Ensemble results of RoBERTa and DistilBERT.} Following \tabref{moreensemble} shows the single and ensemble results of RoBERTa and DistilBERT on the test set
of SEM, TREC, AG, POS, and NER.

\begin{table}[h!]
	\centering
	\renewcommand{\arraystretch}{1.2}
	\resizebox{.4\textwidth}{!}{\begin{tabular}{ccc|ccccr|}
			\cline{4-8}
			                                                  &                                                &        & SEM   & TREC & AG   & POS  & NER   \\ \hline
			\multicolumn{1}{|c|}{\multirow{4}{*}{RoBERTa}}    & \multicolumn{1}{c|}{\multirow{2}{*}{Masking}}  & Single & 11.12 & 3.15 & 5.06 & 2.11 & 11.03 \\
			\multicolumn{1}{|c|}{}                            & \multicolumn{1}{c|}{}                          & Ensem. & 10.54 & 2.40 & 4.55 & 2.11 & 10.57 \\ \cline{2-8}
			\multicolumn{1}{|c|}{}                            & \multicolumn{1}{c|}{\multirow{2}{*}{Finetun.}} & Single & 10.74 & 3.00 & 5.10 & 2.00 & 10.43 \\
			\multicolumn{1}{|c|}{}                            & \multicolumn{1}{c|}{}                          & Ensem. & 10.74 & 2.60 & 4.50 & 1.96 & 9.54  \\ \hline
			\multicolumn{1}{|c|}{\multirow{4}{*}{DistilBERT}} & \multicolumn{1}{c|}{\multirow{2}{*}{Masking}}  & Single & 11.89 & 3.70 & 5.71 & 2.39 & 10.40 \\
			\multicolumn{1}{|c|}{}                            & \multicolumn{1}{c|}{}                          & Ensem. & 11.60 & 3.00 & 5.29 & 2.54 & 9.86  \\ \cline{2-8}
			\multicolumn{1}{|c|}{}                            & \multicolumn{1}{c|}{\multirow{2}{*}{Finetun.}} & Single & 11.94 & 3.30 & 5.42 & 2.39 & 10.18 \\
			\multicolumn{1}{|c|}{}                            & \multicolumn{1}{c|}{}                          & Ensem. & 11.48 & 3.00 & 4.84 & 2.29 & 9.74  \\ \hline
		\end{tabular}}
	\caption{Error rate (\%) on test set of tasks by RoBERTa and DistilBERT.
		Single: the averaged performance of four models with different random seeds.
		Ensem.: ensemble of the four models.}
	\tablabel{moreensemble}
\end{table}

\section{Numerical Values of Plots}
\subsection{Layer-wise behaviors} \seclabel{layerwisenumerical}
\tabref{different_number_of_layers} details the numerical values of \figref{different_layers}.

\begin{table*}[!h]
	\centering
	\resizebox{1.\textwidth}{!}{%
		\begin{tabular}{lccccccccc}
			\toprule
			                                                         & MRPC              & RTE               & CoLA              \\ \midrule
			Finetuning (BERT + classifier)                           & $0.861 \pm 0.008$ & $0.692 \pm 0.027$ & $0.596 \pm 0.015$ \\ \midrule \midrule
			Masking (BERT 00-11 + classifier, initial sparsity 5\%)  & $0.862 \pm 0.015$ & $0.673 \pm 0.036$ & $0.592 \pm 0.004$ \\
			Masking (BERT 00-11 + classifier, initial sparsity 15\%) & $0.825 \pm 0.039$ & $0.626 \pm 0.040$ & $0.522 \pm 0.027$ \\ \midrule \midrule
			Masking (BERT 02-11 + classifier, initial sparsity 5\%)  & $0.868 \pm 0.011$ & $0.695 \pm 0.030$ & $0.595 \pm 0.010$ \\
			Masking (BERT 02-11 + classifier, initial sparsity 15\%) & $0.844 \pm 0.024$ & $0.662 \pm 0.021$ & $0.556 \pm 0.012$ \\ \midrule
			Masking (BERT 04-11 + classifier, initial sparsity 5\%)  & $0.861 \pm 0.004$ & $0.705 \pm 0.037$ & $0.583 \pm 0.005$ \\
			Masking (BERT 04-11 + classifier, initial sparsity 15\%) & $0.861 \pm 0.009$ & $0.669 \pm 0.014$ & $0.553 \pm 0.014$ \\ \midrule
			Masking (BERT 06-11 + classifier, initial sparsity 5\%)  & $0.862 \pm 0.004$ & $0.696 \pm 0.027$ & $0.551 \pm 0.006$ \\
			Masking (BERT 06-11 + classifier, initial sparsity 15\%) & $0.868 \pm 0.008$ & $0.691 \pm 0.033$ & $0.534 \pm 0.016$ \\ \midrule
			Masking (BERT 08-11 + classifier, initial sparsity 5\%)  & $0.848 \pm 0.016$ & $0.675 \pm 0.034$ & $0.538 \pm 0.014$ \\
			Masking (BERT 08-11 + classifier, initial sparsity 15\%) & $0.851 \pm 0.009$ & $0.688 \pm 0.022$ & $0.545 \pm 0.005$ \\ \midrule \midrule
			Masking (BERT 00-09 + classifier, initial sparsity 5\%)  & $0.859 \pm 0.012$ & $0.683 \pm 0.031$ & $0.589 \pm 0.011$ \\
			Masking (BERT 00-09 + classifier, initial sparsity 15\%) & $0.820 \pm 0.052$ & $0.604 \pm 0.021$ & $0.514 \pm 0.016$ \\ \midrule
			Masking (BERT 00-07 + classifier, initial sparsity 5\%)  & $0.829 \pm 0.032$ & $0.649 \pm 0.053$ & $0.574 \pm 0.012$ \\
			Masking (BERT 00-07 + classifier, initial sparsity 15\%) & $0.807 \pm 0.042$ & $0.600 \pm 0.027$ & $0.509 \pm 0.004$ \\ \midrule
			Masking (BERT 00-05 + classifier, initial sparsity 5\%)  & $0.814 \pm 0.033$ & $0.632 \pm 0.058$ & $0.565 \pm 0.027$ \\
			Masking (BERT 00-05 + classifier, initial sparsity 15\%) & $0.781 \pm 0.032$ & $0.567 \pm 0.030$ & $0.510 \pm 0.025$ \\ \midrule
			Masking (BERT 00-03 + classifier, initial sparsity 5\%)  & $0.791 \pm 0.026$ & $0.606 \pm 0.027$ & $0.535 \pm 0.034$ \\
			Masking (BERT 00-03 + classifier, initial sparsity 15\%) & $0.776 \pm 0.035$ & $0.600 \pm 0.019$ & $0.527 \pm 0.014$ \\
			\bottomrule
		\end{tabular}%
	}
	\caption{
		Numerical value of the layer-wise behavior experiment.
		We train for 10 epochs with mini-batch size 32.
		The learning rate is finetuned using the mean results on four different random seeds.
	}
	\tablabel{different_number_of_layers}
\end{table*}

\subsection{Memory consumption}
\tabref{tab:memorycomp} details the numerical values of \figref{memorycomp}.

\begin{table*}[!h]
	\centering
	\renewcommand{\arraystretch}{1.5}
	\resizebox{\textwidth}{!}{
		\begin{tabular}{c|rr|rr|}
			\cline{2-5}
			&\multicolumn{2}{c|}{Number of Parameters} &\multicolumn{2}{c|}{Memory Usage (Kilobytes)} \\ \cline{2-5}
			                           & Finetuning             & Masking                        & Finetuning             & Masking                       \\ \hline
			\multicolumn{1}{|c|}{Pretrained} & \multicolumn{2}{c|}{109,482,240}                              &\multicolumn{2}{c|}{437,928.96}             \\ \hline
			\multicolumn{1}{|c|}{MRPC} & + 1,536                & + 1,536 + 71,368,704 + 1,536   & + 6.144                & + 6.144 + 8,921.088 + 0.192   \\
			\multicolumn{1}{|c|}{SST2} & + 1,536 + 109,482,240  & + 71,368,704 + 1,536           & + 6.144   + 437,928.96 & + 8,921.088 + 0.192           \\
			\multicolumn{1}{|c|}{CoLA} & + 1,536 + 109,482,240  & + 71,368,704 + 1,536           & + 6.144   + 437,928.96 & + 8,921.088 + 0.192           \\
			\multicolumn{1}{|c|}{RTE}  & + 1,536 + 109,482,240  & + 71,368,704 + 1,536           & + 6.144   + 437,928.96 & + 8,921.088 + 0.192           \\
			\multicolumn{1}{|c|}{QNLI} & + 1,536 + 109,482,240  & + 71,368,704 + 1,536           & + 6.144   + 437,928.96 & + 8,921.088 + 0.192           \\
			\multicolumn{1}{|c|}{SEM}  & + 1,536 + 109,482,240  & + 71,368,704 + 1,536           & + 6.144   + 437,928.96 & + 8,921.088 + 0.192           \\
			\multicolumn{1}{|c|}{TREC} & + 4,608 + 109,482,240  & + 4,608  + 71,368,704 + 4,608  & + 18.432  + 437,928.96 & + 18.432 + 8,921.088 + 0.576  \\
			\multicolumn{1}{|c|}{AG}   & + 3,072 + 109,482,240  & + 3,072  + 71,368,704 + 3,072  & + 12.288  + 437,928.96 & + 12.288 + 8,921.088 + 0.384  \\
			\multicolumn{1}{|c|}{POS}  & + 37,632 + 109,482,240 & + 37,632 + 71,368,704 + 37,632 & + 150.528 + 437,928.96 & + 150.528 + 8,921.088 + 4.704 \\
			\multicolumn{1}{|c|}{NER}  & + 6,912  + 109,482,240 & + 6,912  + 71,368,704 + 6,912  & + 27.648  + 437,928.96 & + 27.648  + 8,921.088 + 0.864 \\
			\multicolumn{1}{|c|}{SWAG} & + 768	  + 109,482,240   & + 768    + 71,368,704 + 768    & + 3.072   + 437,928.96 & + 3.072   + 8,921.088 + 0.096 \\ \hline
		\end{tabular}}
	\caption{Model size comparison when applying masking and finetuning. Numbers are based on BERT-base-uncased.
		Note that our masking scheme enables sharing parameters across tasks:
		tasks with the same number of output dimension can use the same classifier layer.}
	\tablabel{tab:memorycomp}
\end{table*}

\end{document}